\documentclass{article}

\PassOptionsToPackage{numbers,sort&compress}{natbib}
\usepackage{report}
\setcitestyle{numbers,square,comma}

\usepackage[utf8]{inputenc}
\usepackage[T1]{fontenc}
\usepackage{hyperref}
\usepackage{url}
\usepackage{booktabs}
\usepackage{amsfonts}
\usepackage{amsmath}
\usepackage{amssymb}
\usepackage{microtype}
\usepackage{graphicx}
\usepackage{xcolor}
\usepackage{multirow}
\usepackage{subcaption}
\usepackage[most]{tcolorbox}
\usepackage{colortbl}
\usepackage{pifont}
\usepackage{capt-of}
\usepackage{needspace}  

\definecolor{njuPurple}{RGB}{220,205,230}
\definecolor{njuPurpleLight}{RGB}{250,245,252}
\newtcolorbox{abstractbox}{
    colback=njuPurpleLight, colframe=njuPurple,
    boxrule=1pt, arc=4mm,
    left=10pt, right=10pt, top=8pt, bottom=8pt,
    opacityback=0.95
}
\renewenvironment{abstract}
  {\begin{abstractbox}\noindent{\large\bfseries Abstract}\par\vspace{4pt}}
  {\end{abstractbox}}

\definecolor{todored}{HTML}{C62828}

\definecolor{macaronpink}{HTML}{FCE4EC}   
\definecolor{macaronmint}{HTML}{E0F2E9}   
\definecolor{macaronlemon}{HTML}{FFF8E1}  
\definecolor{rowtintA}{HTML}{FAFAFA}      
\definecolor{rowtintB}{HTML}{F1F4F8}      

\definecolor{promptboxbg}{HTML}{FDF1F5}      
\definecolor{promptboxframe}{HTML}{E8B4C5}   
\definecolor{promptboxtitlebg}{HTML}{F4C8D5} 
\definecolor{promptboxtitlefg}{HTML}{6E2940} 
\definecolor{configboxbg}{HTML}{F2F9EE}      
\definecolor{configboxframe}{HTML}{B5D6A7}   
\definecolor{configboxtitlebg}{HTML}{CDE6BF} 
\definecolor{configboxtitlefg}{HTML}{2E5A2E} 

\newtcolorbox{promptbox}[1]{%
  enhanced,
  breakable,
  colback=promptboxbg, colframe=promptboxframe,
  boxrule=0.4pt, arc=2.5pt, outer arc=2.5pt,
  left=8pt, right=8pt, top=6pt, bottom=6pt,
  fonttitle=\scriptsize\bfseries\sffamily,
  coltitle=promptboxtitlefg,
  colbacktitle=promptboxtitlebg,
  attach boxed title to top left={xshift=10pt, yshift=-7pt},
  boxed title style={
    colback=promptboxtitlebg, colframe=promptboxtitlebg,
    boxrule=0pt, arc=2pt, outer arc=2pt,
    left=6pt, right=6pt, top=2pt, bottom=2pt,
  },
  title={#1},
  top=10pt,
  fontupper=\scriptsize\ttfamily,
  before skip=8pt, after skip=8pt,
}
\newtcolorbox{promptboxbreak}[1]{%
  enhanced,
  breakable,
  colback=promptboxbg, colframe=promptboxframe,
  boxrule=0.4pt, arc=2.5pt, outer arc=2.5pt,
  left=8pt, right=8pt, top=6pt, bottom=6pt,
  fonttitle=\scriptsize\bfseries\sffamily,
  coltitle=promptboxtitlefg,
  colbacktitle=promptboxtitlebg,
  attach boxed title to top left={xshift=10pt, yshift=-7pt},
  boxed title style={
    colback=promptboxtitlebg, colframe=promptboxtitlebg,
    boxrule=0pt, arc=2pt, outer arc=2pt,
    left=6pt, right=6pt, top=2pt, bottom=2pt,
  },
  title={#1},
  top=10pt,
  fontupper=\scriptsize\ttfamily,
  before skip=8pt, after skip=8pt,
}
\newtcolorbox{configbox}[1]{%
  enhanced,
  breakable,
  colback=configboxbg, colframe=configboxframe,
  boxrule=0.4pt, arc=2.5pt, outer arc=2.5pt,
  left=8pt, right=8pt, top=6pt, bottom=6pt,
  fonttitle=\scriptsize\bfseries\sffamily,
  coltitle=configboxtitlefg,
  colbacktitle=configboxtitlebg,
  attach boxed title to top left={xshift=10pt, yshift=-7pt},
  boxed title style={
    colback=configboxtitlebg, colframe=configboxtitlebg,
    boxrule=0pt, arc=2pt, outer arc=2pt,
    left=6pt, right=6pt, top=2pt, bottom=2pt,
  },
  title={#1},
  top=10pt,
  fontupper=\scriptsize,
  before skip=8pt, after skip=8pt,
}

\title{Solvita: Enhancing Large Language Models for Competitive Programming
via Agentic Evolution}

\author{
\textbf{Han Li}$^{1}$\,\thanks{Equal contribution.}\quad
\textbf{Jinyu Tian}$^{3}$\,\footnotemark[1]\quad
\textbf{Rili Feng}$^{1}$\,\footnotemark[1]\quad
\textbf{Yuqiao Du}$^{2}$\,\footnotemark[1]\quad
\textbf{Chong Zheng}$^{2}$\quad
\textbf{Chenyu Wang}$^{1}$\\
\textbf{Chenchen Liu}$^{3}$\quad
\textbf{Shihao Li}$^{1}$\quad
\textbf{Xinping Lei}$^{1}$\quad
\textbf{Yifan Yao}$^{1}$\quad
\textbf{Weihao Xie}$^{1}$\quad
\textbf{Letian Zhu}$^{1}$\quad
\textbf{Jiaheng Liu}$^{1}$\,\thanks{Corresponding author.}\\[2pt]
\vspace{3mm}
\normalsize
$^{1}$\,\textbf{Nanjing University}\quad
$^{2}$\,\textbf{Tsinghua University}\quad
$^{3}$\,\textbf{Independent Researcher} \\
\vspace{2mm}
    \texttt{han.li.cs@smail.nju.edu.cn},
    \texttt{liujiaheng@nju.edu.cn} \\
}

\date{}

\hypersetup{
	pdftitle={Enhancing Large Language Models for Competitive Programming via Agentic Evolution},
	pdfkeywords={LLM agents, competitive programming, self evolution, adversarial validation, knowledge networks},
}

\begin{document}
\maketitle

\begin{abstract}

Large language models (LLMs) still struggle with the rigorous reasoning demands of hard competitive programming. While recent multi-agent frameworks attempt to bridge this reliability gap, they remain fundamentally stateless: they rely on static retrieval and discard the valuable problem-solving and debugging experience gained from previous tasks. To address this, we present Solvita, an agentic evolution framework that enables continuous learning without requiring weight updates to the underlying LLM.
Solvita reorganizes problem-solving into a closed-loop system of strategy selection, program synthesis, certified supervision, and targeted hacking, executed by four specialized agents (Planner, Solver, Oracle, and Hacker). Crucially, each agent is paired with a trainable, graph-structured knowledge network. As the system operates, outcome signals--such as pass/fail verdicts, test certification quality, and adversarial vulnerabilities discovered by the Hacker--are recast as reinforcement learning updates to these network weights. This allows the agents to dynamically route future queries based on past successes and failures, effectively accumulating transferable reasoning experience over time.
Evaluated across CodeContests, APPS, AetherCode, and live Codeforces rounds, Solvita establishes a new state-of-the-art among code-generation agents, 
outperforming existing multi-agent pipelines and nearly doubling the accuracy of single-pass baselines.

\end{abstract}

\section{Introduction}
\label{sec:intro}

Algorithmic problem-solving requires translating a natural language specification into a correct, efficient program. This demands precise formalization, deliberate strategy selection, and rigorous verification. Competitive programming benchmarks capture these demands in a controlled setting and have become a standard testbed for evaluating structured reasoning in large language models (LLMs) 
~\cite{li2022alphacode,hendrycks2021apps,shi2024can,jain2024livecodebench}.

Despite rapid advancements, the dominant paradigm for LLM coding remains single-shot generation, which conflates understanding, planning, coding, and verification into one monolithic LLM call~\cite{chen2021codex}. Recent multi-step pipelines, such as AlphaCodium~\cite{ridnik2024alphacodium} and MapCoder~\cite{islam2024mapcoder}, mitigate this by introducing hierarchical planning and iterative refinement. Yet, they remain fundamentally stateless: each new problem is solved from scratch, and any experience gained from past mistakes is discarded. Retrieval-augmented generation (RAG) variants~\cite{lewis2020rag} attempt to add memory, but they treat retrieval as a static similarity lookup. Injecting raw text back into a prompt does not fundamentally alter the underlying reasoning procedure. In contrast, strong human programmers improve precisely because they accumulate transferable experience: they learn which strategies fit specific problem structures, recognize why certain implementations tend to fail, and learn to rigorously attack their own solutions before the judge does.


To bridge this gap, we introduce \textbf{Solvita}\footnote{The name \emph{Solvita} combines \emph{solve} with the Latin \emph{vita} (``life''; cf.\ Italian \emph{vita}, English \emph{vital}/\emph{vitality}), evoking a solver that brings life to problem solving.}, a multi-agent framework that brings continuous, experience-driven evolution to LLMs for competitive programming. Solvita replaces static pipelines with a dynamic, closed-loop ecosystem of four specialized agents—a Planner, a Solver, an Oracle, and a Hacker. Rather than finetuning the massive parameters of the underlying LLM, Solvita pairs each agent with a trainable, graph-structured knowledge network. As the system solves problems and generates tests, it uses reinforcement learning to update the routing weights of these networks based on pass/fail verdicts and adversarial discoveries. Consequently, the framework accumulates and reuses experience across a stream of tasks, allowing earlier solving and debugging episodes to directly shape how subsequent problems are approached.


\textbf{(1) An agentic evolution framework}: We reorganize problem-solving into a closed loop of strategy selection (Planner), program synthesis and patch-based repair (Solver), certified internal supervision (Oracle), and targeted adversarial testing (Hacker). This interactive loop ensures that failure signals propagate system-wide, allowing the agents to collaborate and cross-correct.

\textbf{(2) Trainable knowledge networks as macro-level memory}: Instead of a flat document store, each agent is backed by a structured graph linking problem queries, metacognitive analyses, and reusable algorithmic skills. Edge weights are dynamically adjusted by outcome signals rather than static semantic similarity. This transforms memory from passive retrieval into a learned routing mechanism that expands precisely where the agent currently struggles.

\textbf{(3) Agentic feedback as a training signal}: Solvita recasts the outcome signals produced by the Oracle and Hacker—such as certification quality and adversarial vulnerability events—as reinforcement learning signals. The LLM backbone remains entirely frozen, yet the system's reasoning capability improves monotonically with use as the knowledge network learns to route problems to the correct strategies and failure-prevention tactics.

\textbf{(4) 
Excellent performance}:
Solvita establishes a new state-of-the-art among code-generation frameworks on multiple benchmarks. Most notably, with a GPT-5.4 backbone on CodeContests, Solvita lifts pass@1 accuracy from 40.0\% (single-pass) to 82.4\%—nearly doubling the cold-start baseline while maintaining a similar token-consumption footprint to existing multi-agent pipelines.

\section{Data}
\label{sec:data}

\begin{figure*}[t]
	\centering
	\includegraphics[width=0.96\textwidth]{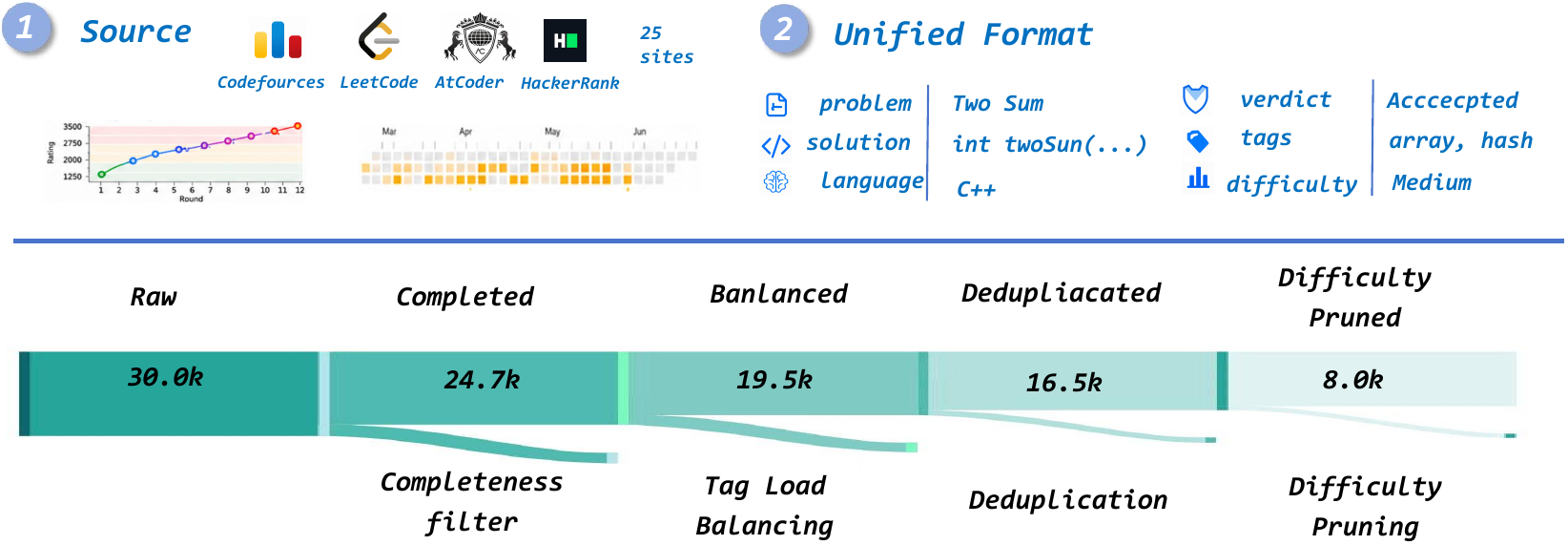}
	\caption{Data pipeline overview. Raw artifacts are collected from multiple competitive programming platforms, normalized into a unified JSON schema, and passed through four filtering stages: completeness, tag load balancing, embedding deduplication, and per tag difficulty pruning.}
	\label{fig:data_pipeline}
\end{figure*}

Each knowledge network requires a cold-start corpus before online learning can begin. We build this corpus in three steps---collection from heterogeneous competitive-programming sources, schema unification into a single JSON record format, and a four-stage filtering pipeline that controls completeness, tag balance, redundancy, and difficulty.

\subsection{Collection}
\label{sec:data:collection}

Our raw corpus is gathered directly from major competitive-programming judges---Codeforces, AtCoder, Aizu Online Judge, and a long tail of smaller platforms (e.g., LeetCode, SPOJ, UOJ)---rather than from any single pre-packaged dataset. Where applicable we cross-check against existing public collections such as CodeContests~\citep{li2022alphacode}, CodeContests+~\citep{wang2025codecontests}, and APPS~\citep{hendrycks2021apps} to recover editorial text and verdict labels that the raw scrape misses, but the unit of collection is the platform, not the dataset. The corpus is split into two subsets. The \emph{training subset} keeps as much associated information as each source exposes, including statements, tests, metadata, editorials, and submissions with verdicts, and feeds the agentic-evolution pipeline. The \emph{skill subset} pairs canonical reference problems with their official editorial solutions, seeding the downstream skill library.
\subsection{Schema Unification}
\label{sec:data:unification}

To reconcile the heterogeneous formats used across platforms, we normalize all collected artifacts into a unified JSON schema. Each record exposes a fixed set of canonical fields covering the problem statement, typed variable declarations and constraint bounds, sample and hidden tests stored as I/O pairs, editorial text, submission source with judge verdict and execution time, and algorithmic tags drawn from a controlled vocabulary. Interactive protocols, special judges, and multi-test-case packing conventions are mapped to standardized flags. The unified schema is the single input format consumed by all downstream stages.

\subsection{Filtering}
\label{sec:data:filtering}

Starting from 30,018 problems, the unified corpus is refined through four sequential filters. The order is deliberate: tag load balancing precedes deduplication so that the expensive embedding-based dedup operates on a substantially smaller per-tag bucket, and difficulty pruning is applied last so that the floor is set against the post-dedup distribution rather than its inflated raw counterpart. \ding{182}~\textbf{Completeness.} A problem is retained only if it carries both public and private test sets, algorithmic tags, a difficulty signal (e.g., rating, tier, or division), and a parseable I/O specification with explicit constraint bounds; this stage leaves 24,712 problems. \ding{183}~\textbf{Tag load balancing.} A handful of tags would otherwise dominate the corpus and bias training toward their solution patterns, so we cap each tag at $C_{\max}$ problems and subsample the over represented tags, keeping the per tag distribution within a constant factor of the smallest tag and yielding 19,486 problems. \ding{184}~\textbf{Deduplication.} We embed every statement with \texttt{text-embedding-3-large}, bucket by algorithmic tag, and compute pairwise cosine similarity within each bucket; any pair with similarity above $\delta$ is marked duplicate, and we retain the variant with more attached submissions and a larger test set, leaving 16,503 problems. \ding{185}~\textbf{Difficulty pruning.} Trivially easy problems are removed because the base LLM solves them without experience augmentation: survivors are ranked by platform difficulty and any problem below the per tag floor $d_{\min}^{(\tau)}$ is dropped, producing the final corpus of 8,017 problems. The per platform difficulty mapping, the value of $\delta$, the per tag floors $d_{\min}^{(\tau)}$, the cap $C_{\max}$, and the per platform raw and surviving counts are tabulated in Appendix~\ref{app:data:pipeline}.

\section{Solvita}
\label{sec:method}

\subsection{Framework Overview}
\label{sec:method:arch}

\begin{figure*}[t]
	\centering
	\includegraphics[width=\textwidth]{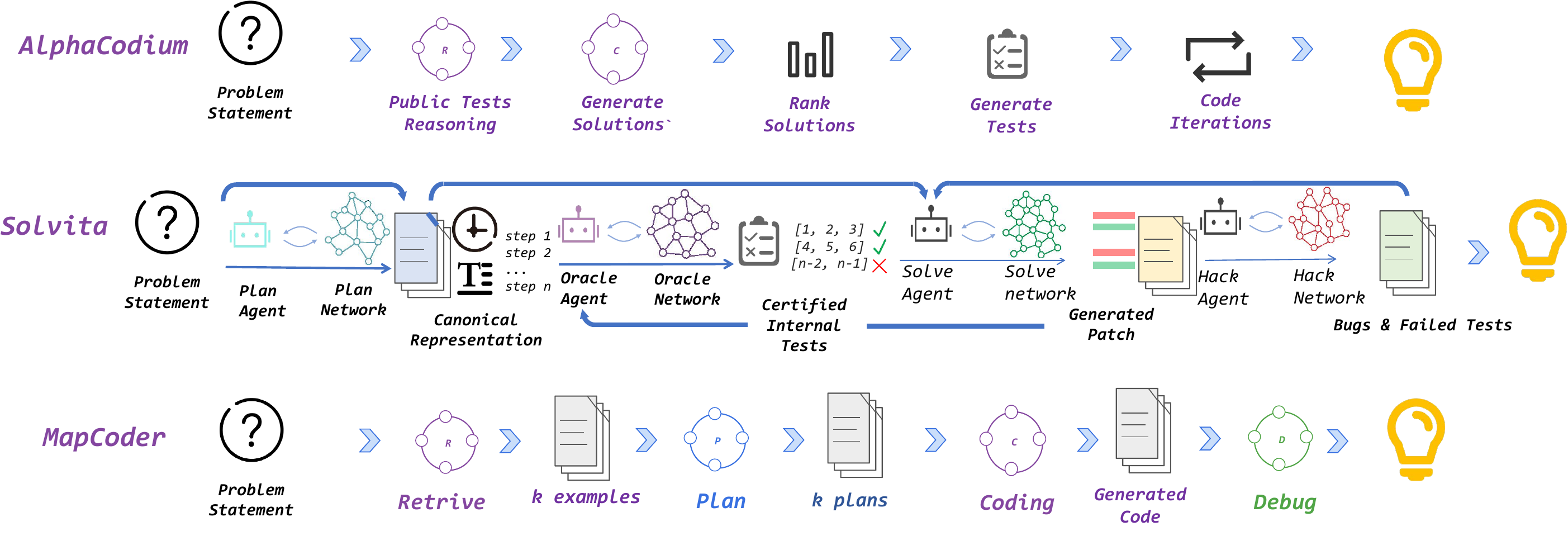}
	\caption{The Solvita architecture and its comparison with existing agent frameworks. Solvita couples an Oracle for certified internal-test construction, a Planner--Solver loop with patch-based refinement, and a Hacker that launches cascading adversarial attacks; failure signals propagate across all four agents' knowledge networks (dashed arrows). In contrast to prior single-agent or pipeline-style code agents, Solvita closes the solve--certify--attack loop within one shared knowledge substrate.}
	\label{fig:pipeline}
\end{figure*}

To this end, \textbf{Solvita} is built around four cooperating agents---a \textbf{Planner} that canonicalizes the problem and selects a paradigm, a \textbf{Solver} that implements the strategy and repairs it via search-and-replace patches rather than full regeneration, an \textbf{Oracle} that builds a certified internal test suite, and a \textbf{Hacker} that mounts adversarial attacks---coupled into one closed loop in which any failure signal propagates across all four agents (Figure~\ref{fig:pipeline}). Since interaction signals vary widely in their informativeness, each agent is backed by its own trainable knowledge network under a role-specific schema. These networks share a common contextual-bandit policy~\cite{li2010contextual}, which surfaces the most informative precedents as advisory context at inference time and retires entries whose running reward falls below a deprecation threshold. Full policy details, featurization, and hyperparameters are deferred to Appendix~\ref{app:bandit}. Throughout the paper we use the single term \emph{knowledge network} for these per-agent stores; the remainder of this section describes each agent in turn (Planner, Solver, Oracle, Hacker), together with the knowledge network and reward signal that trains it.

\subsection{Planner}
\label{sec:method:planner}

The Planner first reformulates the raw problem as a purely formal mathematical specification, stripping narrative framing and problem-irrelevant context to expose the underlying objective, variables, and constraints. From this canonical form it proposes a strategy: a predicted set of algorithmic tags, an implementation sketch, and a complexity estimate. On replanning after failure, the classified failure verdict guides revision. The exact \texttt{abstract\_problem} prompt and JSON output schema are listed in Appendix~\ref{app:prompts}. The knowledge network behind the Planner stores strategy records linking the formalized problem to its predicted tags and downstream outcome, and at inference time the bandit policy of Section~\ref{sec:method:arch} retrieves precedents from structurally similar formalizations as planning advice.

\paragraph{Training.}
The Planner knowledge network is trained with the shared bandit update of Section~\ref{sec:method:arch} under a tag-prediction reward against the problem's ground-truth tag set: each predicted tag earns $r = +1$ if it matches a true tag and $r = -1$ otherwise, summed over the prediction. This directly teaches the network which formalized problem structures admit, which algorithmic paradigms.

\subsection{Solver}
\label{sec:method:solver}

The Solver implements the selected strategy as a C++ program, self-validating against public and Oracle-generated tests. On subsequent iterations, it applies \emph{patch-based repair}: rather than regenerating the full solution, it emits search-and-replace edit blocks targeting only the diagnosed fault. A patch is accepted only if all regression tests (previously passing) still pass, preserving prior correctness while focusing effort on the unresolved case. The full prompt set (initial generation, patch decision, SEARCH/REPLACE patch, regeneration, and failure analysis) is given in Appendix~\ref{app:prompts}, and the storage layout, featurizer, and event-propagation mechanism for the Solver's three-layer query--method--skill (QMS) knowledge network are documented in Appendix~\ref{app:examples}.

\begin{figure*}[t]
	\centering
	\includegraphics[width=0.9\textwidth]{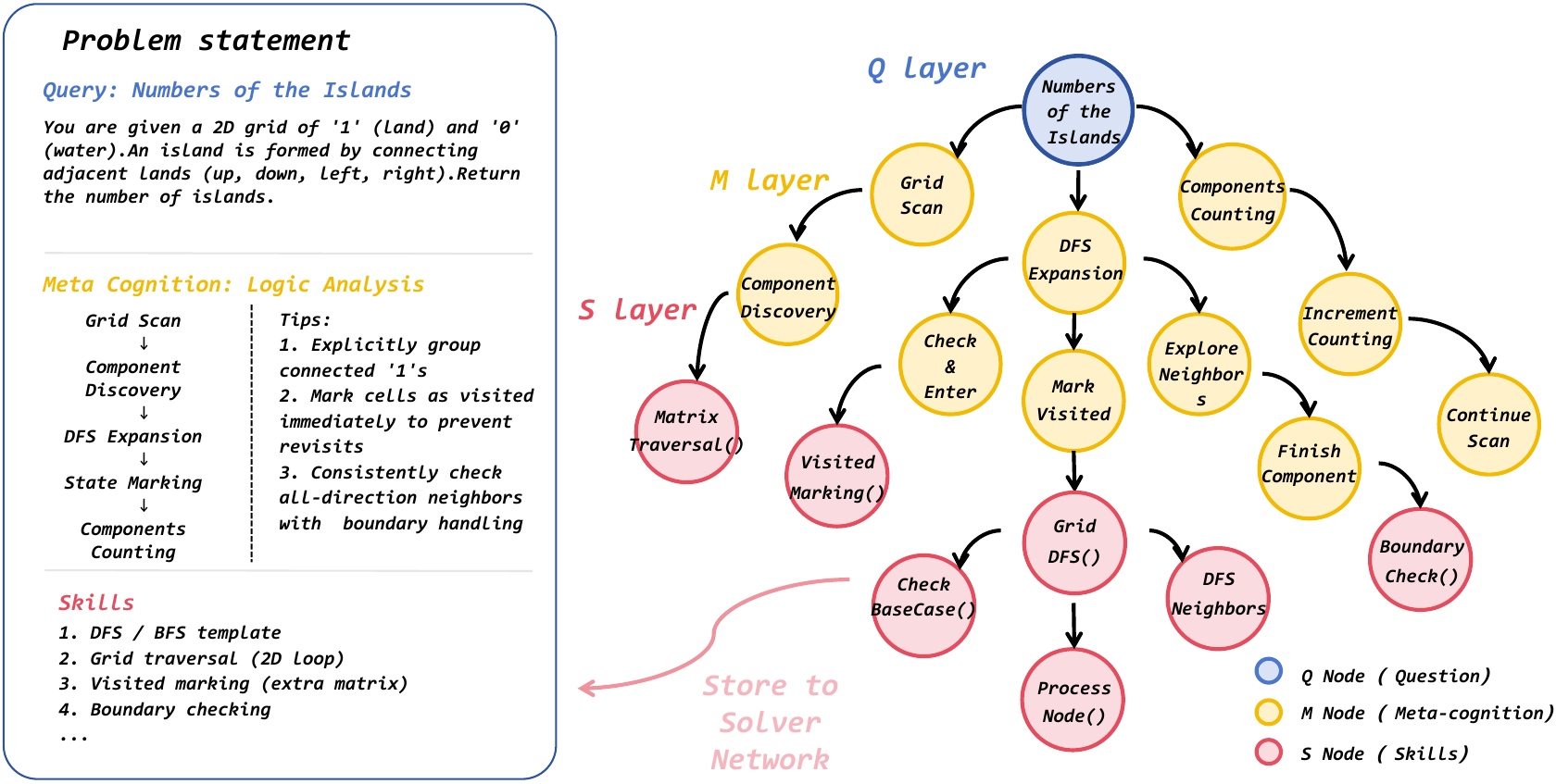}
	\caption{The three-layer Solver knowledge network. Q nodes (top) store problem descriptions and metadata; M nodes (middle) hold contrastive or analysis solution decompositions as function-block DAGs; S nodes (bottom) store annotated skills with code templates. Solid arrows denote deterministic links via function-block identifiers; dashed arrows denote semantic-similarity fallback. Edge thickness reflects learned weight magnitude.}
	\label{fig:solver_network}
\end{figure*}

The Solver is backed by a three-layer heterogeneous directed graph $\mathcal{G} = (\mathcal{V}_Q \cup \mathcal{V}_M \cup \mathcal{V}_S,\; \mathcal{E}_{QM} \cup \mathcal{E}_{MS})$ (Figure~\ref{fig:solver_network}). \textbf{Q nodes} store the description and metadata of previously encountered problems. \textbf{M nodes} decompose solutions into function-block DAGs; \emph{contrastive} M nodes pair a correct and incorrect solution sharing the same approach to isolate failure points, while \emph{analysis} M nodes summarize accepted solutions as standalone trajectories. \textbf{S nodes} store annotated algorithmic skills with C++ templates, linked to M nodes via function-block identifiers (deterministic match or embedding fallback). Given a new problem $q_{\text{new}}$, the top-$k$ similar Q nodes are retrieved and expanded into an activated subgraph; each reachable skill $s$ receives a selection score aggregated over all two-hop paths,
\begin{equation}
	\rho(s \mid q_{\text{new}})
	\;=\!\!\sum_{\substack{q_i,\, m_j \,:\\ q_i \to m_j \to s \,\in\, \mathcal{G}}}\!\!
	\operatorname{Sim}\!\bigl(q_{\text{new}},\, q_i\bigr)
	\,\cdot\, w_{\mathrm{qm}}^{(i,j)}
	\,\cdot\, w_{\mathrm{ms}}^{(j,s)},
	\label{eq:path_prob}
\end{equation}
where $w_{\mathrm{qm}}$ and $w_{\mathrm{ms}}$ are learned edge weights. Skills are sampled from $\pi(s) = \operatorname{softmax}(\rho(s)/T)$ and assembled with their associated problem descriptions and contrastive analyses into a structured augmentation block.

\paragraph{Training via contrastive REINFORCE.}
The Solver's training optimizes the Solver knowledge network without modifying the frozen LLM backbone. For each training problem, the agent solves it twice with the same backbone: once conditioned on the Solver knowledge network (full skill augmentation) and once without it (bare LLM). The outcome difference $\Delta R = R_{\text{with}} - R_{\text{without}}$ serves as a counterfactual reward isolating the network's contribution, where $R$ is the test pass rate. Edge weights are updated by REINFORCE:
\begin{equation}
	\nabla_{\mathbf{w}} \mathcal{L}
	\;=\; -\,\Delta R \,\cdot\, \nabla_{\mathbf{w}} \log p\bigl(\mathbf{s} \,\big|\, q_{\text{new}}\bigr),
	\qquad
	\mathbf{w} \,\leftarrow\, \mathbf{w} \,-\, \alpha \,\cdot\, \nabla_{\mathbf{w}} \mathcal{L},
	\label{eq:rl-loss}
\end{equation}
where the gradient decomposes through the chain rule from skill probabilities $\pi$ through $\rho$ to the underlying QM and MS edge weights, with MS weight groups renormalized after each update. The graph also grows dynamically: when both variants succeed, no node is added; when both fail, a new contrastive M node pairs the incorrect output with the closest correct solution from the corpus; when outcomes differ, the correct and incorrect outputs are directly paired. This ensures that the graph expands precisely where the agent currently struggles.

\subsection{Category-Aligned Strategy Taxonomy of Oracle and Hacker Memory}
\label{sec:method:strategy_taxonomy}

\begin{figure*}[t]
    \centering
    \includegraphics[width=\textwidth]{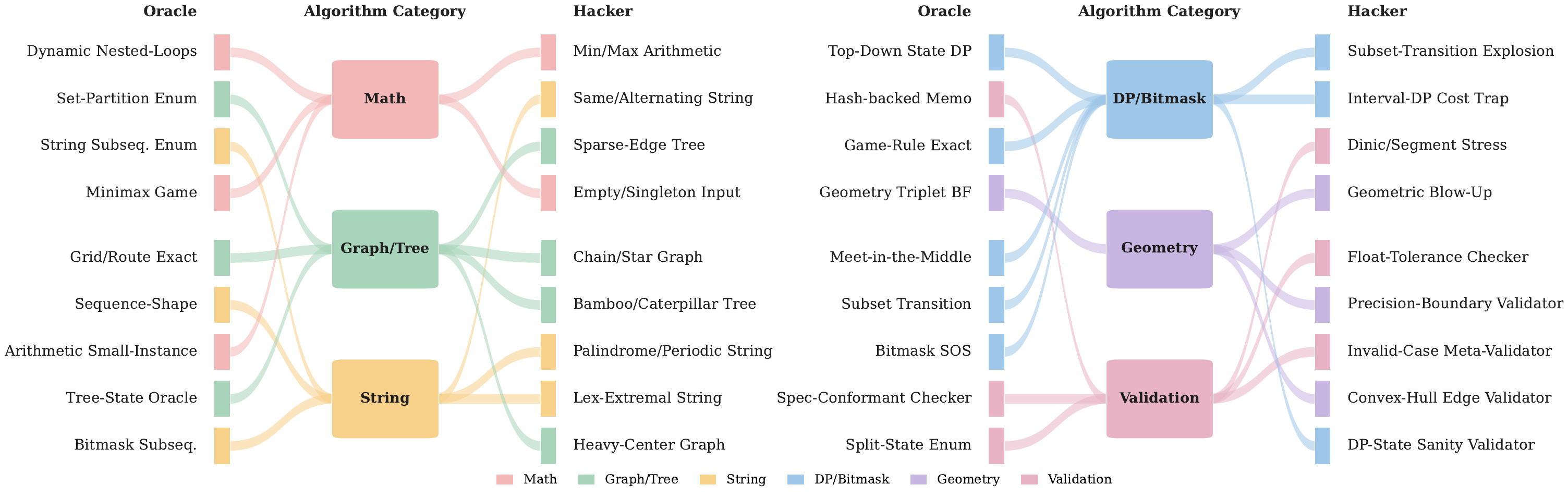}
    \caption{Seed-level strategy taxonomy of Oracle and Hacker memory, showing how each agent factorizes the shared algorithm space into its own reusable strategy units.}
    \label{fig:strategy_taxonomy}
\end{figure*}

Before specifying how the Oracle and Hacker are individually trained, we describe the shared structure their knowledge networks converge to and why that structure is functionally complementary. Figure~\ref{fig:strategy_taxonomy} presents the seed-level view: the center column lists common competitive-programming categories, and the left and right columns show how Oracle and Hacker decompose that same space into different strategy families and representative reusable seeds.

The two sides overlap on categories but factor the same problem space in functionally different ways. Oracle strategies concentrate on routes that produce reliable supervision---DP/Search and Enumeration families dominate, with Decomposition and Domain-Aware as secondary modes---and align primarily with categories where reference solvers and cross-checkers are most informative (DP, Graph, Math, Bitmask, String). Hacker strategies, in contrast, concentrate on routes that expose latent bugs---Complexity/Worst-case and Structural/Topology dominate, with Boundary/Corner and Checker/Validation as secondary modes---and align with categories where stress testing and validator design carry the most signal (Stress, Checker, Graph, DP, String). Within each family the seeds are internally heterogeneous rather than collapsing to a single heuristic, with full counts and per-category percentages reported in Figure~\ref{fig:strategy_taxonomy}. This motivates Solvita's design choice of \emph{trainable} role-specific knowledge networks and the two distinct reward functions developed in Sections~\ref{sec:method:oracle} and \ref{sec:method:hack}: instead of retrieving raw past examples, the system learns to route each new problem toward the most suitable certification strategy and the most informative adversarial-testing strategy.

\subsection{Oracle}
\label{sec:method:oracle}

The Oracle produces certified supervision for the pipeline through four stages: (1)~generate a testlib-based C++ generator, validator, and optional custom checker with iterative self-repair; (2)~verify that the reference solver reproduces all public sample outputs; (3)~generate $N_{\text{target}}$ additional test inputs and certify each against an independent judge (custom checker $>$ correct-solution runner $>$ exact match), yielding a certification ratio $\rho = N_{\text{cert}} / N_{\text{target}} \in [0,1]$; and (4)~accept the artifact only after checking that the generated input set $I$ and expected-output set $O$ are nonempty, the certification ratio clears the threshold $\tau$, and multi-answer routes provide a nonempty custom-checker evidence set $C_{\mathrm{ma}}$:
\begin{equation}
	\begin{split}
	A(x,f) \;=\; \mathbb{1}\bigl[\,& |I| > 0 \,\wedge\, |O| > 0 \,\wedge\, \rho \ge \tau \\
	&\wedge\, \bigl(\mathrm{route}(x) \neq \text{multi\_answer} \,\vee\, C_{\mathrm{ma}} \neq \emptyset\bigr)\bigr].
	\end{split}
	\label{eq:oracle_gate}
\end{equation}
Artifacts failing the gate are discarded, and the Oracle retries with an alternative solver family. The Oracle is backed by a network of reference-solver strategy families $\mathcal{F}$ (e.g., top-down DP, constructive enumeration, brute-force verification), each annotated with applicable problem structures and historical success rates; the bandit policy of Section~\ref{sec:method:arch} selects the best-scoring family for each problem's structural context. The four sub-prompts (generator, validator, checker, solver) and the stage-conditioned solver guidance appear in Appendix~\ref{app:prompts}.

\paragraph{Training.}
For each training problem $x$ the Oracle picks a family $f \in \mathcal{F}$ and runs the four-stage pipeline, producing a certification ratio $\rho(x,f) \in [0, 1]$ and, when $\rho = 1$, an external judge verdict. The reward $r_{\text{oracle}}(x, f) \in [-1, +1]$ is the sum of (i) a partial-credit term proportional to $\rho$ when $\rho<1$, (ii) a full-certification bonus signed by the judge verdict when $\rho=1$, and (iii) a failure penalty selected from four mutually exclusive failure modes (no failure / unready state / self-check failure / crash); the exact term coefficients, verdict scores, and penalty values are reported in Appendix~\ref{app:oracle}. The selected family is then updated by the bandit rule of Section~\ref{sec:method:arch} on the active feature keys $\Phi(x)$ for $x$ (problem tags, constraint regime, paradigm class), so the resulting policy steadily concentrates on the family that yields the highest certified-test mass on each problem type.

\subsection{Hacker}
\label{sec:method:hack}

The Hacker searches for adversarial inputs that expose bugs surviving Oracle certification. A code analyst inspects the candidate via sandboxed execution and produces a structured vulnerability report (suspected bug class, attack hypothesis, recommended route). A cascading router then selects an attack route $u \in \mathcal{U} = \{\text{semantic},\, \text{stress},\, \text{antihash}\}$, instantiating respectively corner-case construction, maximum-bound stress testing, and lattice-based hash-collision generation. If the selected route fails, the system cascades through a fallback chain before declaring the candidate safe. The Hacker is backed by a vulnerability catalog recording exploitation types, triggering input patterns, successful attack routes, and algorithmic context; when a bug is discovered, the failure event propagates to all four knowledge networks, so each lesson is internalized once and reused everywhere. The Code-Analyst controller prompt and the three route-specific generator prompts (with their checklist/patch repair variants) are listed in Appendix~\ref{app:prompts}.

\paragraph{Training.}
For each candidate solution and chosen route $u$ the Hacker runs one round, producing a verdict set $V$; if no break is found, the cascade advances to the next route, up to a per-candidate budget of $\texttt{max\_hack\_rounds}$ rounds (default $3$, see Appendix~\ref{app:config} and the sensitivity sweep in Appendix~\ref{app:ablations}). Letting $V_{\text{valid}} \subseteq V$ be the inputs that pass the validator and $V_{\text{break}} \subseteq V_{\text{valid}}$ those that expose a failure, we summarize the round by three signals---the valid-input rate $g_{\text{valid}} = |V_{\text{valid}}| / |V|$, the break rate $g_{\text{break}} = |V_{\text{break}}| / \max(|V_{\text{valid}}|, 1)$, and an average severity $g_{\text{sev}}$ over the broken inputs---and combine them, minus a compile-failure penalty, into
\begin{equation}
	r_{\text{hack}}(x, u) \;=\;
	\operatorname{clip}_{[-1,\,+1]}\!\Bigl(\,
		w_v\,g_{\text{valid}} \,+\, w_b\,g_{\text{break}} \,+\, w_s\,g_{\text{sev}} \,-\, \kappa(c)
	\,\Bigr).
	\label{eq:hack_reward}
\end{equation}
The heaviest weight sits on actually breaking the candidate while still rewarding routes that produce valid, severity-graded inputs; the precise weights $(w_v, w_b, w_s)$, the per-verdict severity table $\omega(\cdot)$, the compile penalty $\kappa(c)$, and the degenerate-round correction for $|V_{\text{valid}}| = 0$ are listed in Appendix~\ref{app:hack}. The selected route is updated by the same bandit rule as the Oracle on the active feature keys $\Phi(x)$, and any successful-break event additionally writes a contrastive entry into the Planner, Solver, and Oracle knowledge networks via the shared event bus, so the Hacker's discoveries directly reshape the other three policies.

\section{Experiments}
\label{sec:exp}

We evaluate Solvita on three competitive-programming benchmarks --- CodeContests~\cite{li2022alphacode} (CC, 165 problems), APPS~\cite{hendrycks2021apps} (1{,}000 sampled across tiers), and AetherCode~\cite{aethercode} (AC, 400 problems) --- and on recent Codeforces rounds. The main comparison uses five frontier backbones (GPT-5.4~\cite{openai2023gpt4}, Claude Opus 4.6~\cite{anthropic2024claude}, Qwen3.6, DeepSeek V4 Pro, Grok), with the same model powering every agent within a run; the more expensive diagnostic and component ablations report representative backbone subsets, always matched within each table. Pass@1 is the primary metric. We compare against commercial coding agents (Codex CLI, Claude Code), open-source agent frameworks (AlphaCodium~\cite{ridnik2024alphacodium}, MapCoder~\cite{islam2024mapcoder}, AgentCoder~\cite{huang2024agentcoder}), and a stateless single-pass generator. Decoding settings, pipeline budgets, and knowledge-network defaults are matched across methods; full configuration is in Appendix~\ref{app:config} and verbatim prompts in Appendix~\ref{app:prompts}.

\subsection{Comparison with Code-Generation Agents}
\label{sec:exp:main}

Solvita attains the best pass@1 in 14 of the 15 backbone--benchmark cells of Table~\ref{tab:main}, with the only exception on AetherCode under Claude Opus 4.6. The two commercial agents are each strongest on their home backbone but lose ground when moved to other models, while the open-source frameworks trail Solvita on every cell, with the gap widening on the harder AetherCode set. The lead grows on stronger backbones, indicating that knowledge accumulation and adversarial validation compound on top of an already capable solver rather than substituting for raw capability.

\begin{table*}[ht]
	\caption{Main results (pass@1, \%) on CodeContests (CC), APPS, and AetherCode (AC). Each backbone is a column group; \textbf{bold} marks the best per column.}
	\label{tab:main}
	\centering
	\scriptsize
	\setlength{\tabcolsep}{2.2pt}
	\renewcommand{\arraystretch}{1.18}
	\resizebox{0.96\linewidth}{!}{%
	\begin{tabular}{l ccc ccc ccc ccc ccc}
		\toprule
		& \multicolumn{3}{c}{\textbf{GPT-5.4}}
		& \multicolumn{3}{c}{\textbf{Claude Opus 4.6}}
		& \multicolumn{3}{c}{\textbf{Qwen3.6}}
		& \multicolumn{3}{c}{\textbf{DeepSeek V4 Pro}}
		& \multicolumn{3}{c}{\textbf{Grok}} \\
		\cmidrule(lr){2-4}\cmidrule(lr){5-7}\cmidrule(lr){8-10}\cmidrule(lr){11-13}\cmidrule(lr){14-16}
		Method & CC & APPS & AC & CC & APPS & AC & CC & APPS & AC & CC & APPS & AC & CC & APPS & AC \\
		\midrule
		Single-pass  & 40.00 & 37.90 & 18.00 & 44.85 & 40.10 & 22.75 & 33.94 & 30.80 & 9.50 & 47.27 & 42.50 & 24.00 & 38.18 & 34.60 & 15.50 \\
		\rowcolor{macaronpink} Codex CLI    & 81.82 & 67.10 & 48.50 & 70.30 & 59.80 & 44.25 & 60.00 & 47.90 & 22.50 & 79.39 & 66.30 & 45.00 & 73.33 & 57.20 & 31.50 \\
		\rowcolor{macaronpink} Claude Code  & 70.91 & 60.40 & 42.75 & 80.00 & 69.00 & \textbf{54.25} & 58.79 & 46.30 & 21.50 & 76.97 & 64.50 & 43.75 & 72.12 & 55.50 & 29.75 \\
		\rowcolor{macaronmint} AlphaCodium  & 60.61 & 52.40 & 33.00 & 64.24 & 56.20 & 36.50 & 53.33 & 42.30 & 14.50 & 70.91 & 53.40 & 33.00 & 60.61 & 46.10 & 21.00 \\
		\rowcolor{macaronmint} MapCoder     & 57.58 & 54.30 & 30.50 & 60.00 & 55.10 & 38.25 & 50.91 & 41.00 & 16.00 & 66.67 & 54.80 & 35.00 & 62.42 & 44.20 & 19.50 \\
		\rowcolor{macaronlemon} \textbf{Solvita (ours)}
			& \textbf{82.42} & \textbf{67.70} & \textbf{49.25}
			& \textbf{80.61} & \textbf{69.30} & 53.75
			& \textbf{69.70} & \textbf{55.10} & \textbf{26.00}
			& \textbf{89.09} & \textbf{68.10} & \textbf{51.50}
			& \textbf{78.18} & \textbf{58.50} & \textbf{33.50} \\
		\bottomrule
	\end{tabular}}
\end{table*}

\begin{figure*}[t]
	\centering
	\begin{subfigure}[t]{0.49\textwidth}
		\centering
		\includegraphics[width=\linewidth]{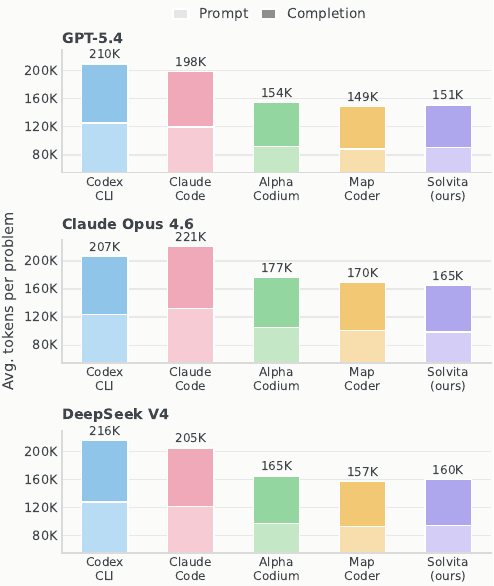}
		\caption{Average token consumption.}
		\label{fig:token_consumption}
	\end{subfigure}
	\hfill
	\begin{subfigure}[t]{0.49\textwidth}
		\centering
		\includegraphics[width=\linewidth]{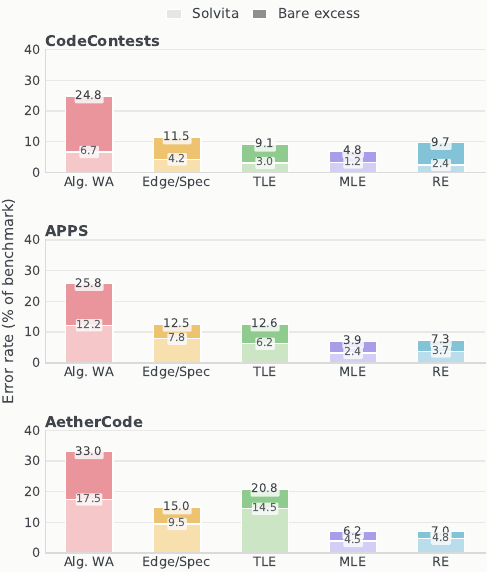}
		\caption{Residual error categories.}
		\label{fig:error_categories}
	\end{subfigure}
	\caption{Cost and failure-profile analysis. (a) Average prompt and completion token consumption per problem, grouped by backbone and agent framework. Each bar stacks prompt and completion tokens. (b) Error-rate decomposition by benchmark and failure type under the representative GPT-5.4 backbone. The light segment shows Solvita's residual error rate, and the darker segment (``Bare excess'') shows the additional error mass of the bare single-pass model; each bar top matches the GPT-5.4 bare-model error rate in Table~\ref{tab:main}. Error categories are defined as follows: Alg.\ WA (Algorithmic Wrong Answer), Edge/Spec (Edge-case or Specification mismatch), TLE (Time Limit Exceeded), MLE (Memory Limit Exceeded), and RE (Runtime Error).}
	\label{fig:cost_error_profile}
\end{figure*}

Beyond pass@1, we also inspect the cost and failure profile of the main comparison. Figure~\ref{fig:cost_error_profile} shows that Solvita stays in the same token-consumption band as open-source agent frameworks rather than matching the substantially higher footprint of commercial CLI agents. The residual-failure decomposition further shows that the gain is not concentrated in a single easy category: relative to the bare single-pass model, Solvita reduces algorithmic, specification-level, complexity, memory, and runtime failures across all three benchmarks.

\subsection{Component and Knowledge-Network Ablation}
\label{sec:exp:ablation}

Table~\ref{tab:ablation} decomposes the contribution of each pipeline element. To establish clear baselines, we define \textbf{Single-pass} as a standard monolithic LLM generation without any multi-agent loop or persistent memory, and \textbf{without training} as the complete multi-agent framework running statelessly (i.e., with empty, untrained knowledge networks). These are compared against the \textbf{Full system}, where all four agents utilize their fully trained knowledge networks after the entire training trajectory. Switching from a single-pass generator to the multi-agent architecture without any persistent network already closes most of the gap to the full system, confirming the value of the closed-loop scaffold. Each trainable network is then probed at three checkpoints along the 5{,}318-problem training trajectory, taken at $1.5\mathrm{k}$, $3\mathrm{k}$, and $4.5\mathrm{k}$ processed problems. Among the three knowledge networks, the Solver knowledge network gives the largest single-component gain on every benchmark, while the Hacker and Oracle knowledge networks add smaller but consistent margins; reading the same network's three rows shows that gains genuinely accumulate with experience rather than appearing all at once. The full system surpasses any single-network addition on every backbone and at every checkpoint, so the components compound rather than substitute.

\begin{table*}[ht]
	\caption{Additive ablation of pipeline components on CC, APPS, and AetherCode (pass@1, \%). Each trainable network is reported at three checkpoints along the 5{,}318-problem training trajectory ($@1.5\mathrm{k}$, $@3\mathrm{k}$, $@4.5\mathrm{k}$ processed problems); static configurations are reported once. (Note: \emph{Single-pass} is a monolithic baseline; \emph{without training} is the stateless multi-agent framework; \emph{Full system} includes fully trained networks).}
	\label{tab:ablation}
	\centering
	\footnotesize
	\setlength{\tabcolsep}{4pt}
	\renewcommand{\arraystretch}{1.0}
	\begin{tabular}{l ccc ccc ccc}
		\toprule
		& \multicolumn{3}{c}{\textbf{GPT-5.4}}
		& \multicolumn{3}{c}{\textbf{Claude Opus 4.6}}
		& \multicolumn{3}{c}{\textbf{Qwen3.6}} \\
		\cmidrule(lr){2-4}\cmidrule(lr){5-7}\cmidrule(lr){8-10}
		Configuration & CC & APPS & AC & CC & APPS & AC & CC & APPS & AC \\
		\midrule
		Single-pass               & 40.00 & 37.90 & 18.00 & 44.85 & 40.10 & 22.75 & 33.94 & 30.80 & 9.50 \\
		\rowcolor{rowtintB} without training          & 67.70 & 54.50 & 35.00 & 66.00 & 55.50 & 39.50 & 55.50 & 42.00 & 12.00 \\
		\midrule
		\rowcolor{rowtintA} \;\;+ Solver network @$1.5\mathrm{k}$ & 70.86 & 57.42 & 38.40 & 69.17 & 58.58 & 42.82 & 58.50 & 44.96 & 15.20 \\
		\rowcolor{rowtintA} \;\;+ Solver network @$3\mathrm{k}$     & 73.63 & 60.18 & 41.38 & 72.05 & 61.28 & 45.78 & 61.13 & 47.55 & 18.00 \\
		\rowcolor{rowtintA} \;\;+ Solver network @$4.5\mathrm{k}$  & 75.60 & 61.80 & 43.50 & 73.80 & 63.20 & 47.80 & 63.00 & 49.40 & 20.00 \\
		\midrule
		\rowcolor{rowtintB} \;\;+ Hacker network @$1.5\mathrm{k}$ & 69.42 & 55.90 & 36.40 & 67.80 & 56.82 & 40.90 & 57.10 & 43.44 & 13.40 \\
		\rowcolor{rowtintB} \;\;+ Hacker network @$3\mathrm{k}$     & 70.93 & 57.13 & 37.63 & 69.38 & 58.00 & 42.13 & 58.50 & 44.70 & 14.63 \\
		\rowcolor{rowtintB} \;\;+ Hacker network @$4.5\mathrm{k}$  & 72.00 & 58.00 & 38.50 & 70.50 & 58.80 & 43.00 & 59.50 & 45.60 & 15.50 \\
		\midrule
		\rowcolor{rowtintA} \;\;+ Oracle network @$1.5\mathrm{k}$ & 70.26 & 56.90 & 38.08 & 68.76 & 57.90 & 42.30 & 58.02 & 44.40 & 14.60 \\
		\rowcolor{rowtintA} \;\;+ Oracle network @$3\mathrm{k}$     & 72.50 & 58.95 & 40.78 & 71.18 & 60.00 & 44.75 & 60.18 & 46.50 & 16.88 \\
		\rowcolor{rowtintA} \;\;+ Oracle network @$4.5\mathrm{k}$  & 74.10 & 60.50 & 42.70 & 72.90 & 61.50 & 46.50 & 61.80 & 48.00 & 18.50 \\
		\midrule
		\rowcolor{macaronlemon} \textbf{Full system}
			& \textbf{82.42} & \textbf{67.70} & \textbf{49.25}
			& \textbf{80.61} & \textbf{69.30} & \textbf{53.75}
			& \textbf{69.70} & \textbf{55.10} & \textbf{26.00} \\
		\bottomrule
	\end{tabular}
\end{table*}

\subsection{Patch-Based Repair vs.\ Full Regeneration}
\label{sec:exp:patch}

Table~\ref{tab:patch} compares the Solver's patch-based inner loop against full regeneration under matched iteration budgets ($N_{\max}=8$, identical retrieval, identical decoding). \emph{Solve} is pass@1; \emph{Iters} is the mean number of Solver iterations actually executed per problem (early-exit allowed once all tests pass); \emph{TokSv} is the relative completion-token saving against a fixed reference cost defined as ``each iteration emits a full solution from scratch and the run consumes the entire iteration budget,'' i.e.~$T_{\text{ref}} = N_{\max}\cdot \bar t_{\text{full}}$, where $\bar t_{\text{full}}$ is the average completion-token count of one full-regeneration draft on that benchmark; saving is then $1 - T_{\text{actual}}/T_{\text{ref}}$. Under this common reference, both strategies show some saving (full regeneration mostly through early-exit) and patch repair shows substantially more, because each post-draft iteration emits only a SEARCH/REPLACE block rather than a fresh solution. In this ablation, patch repair also attains a higher solve rate on every reported benchmark and backbone while running fewer iterations: regeneration plateaus several points lower because each retry rewrites the candidate from scratch and routinely breaks invariants the previous draft already satisfied. The remaining ablation protocol---LLM-skill selection and contrastive vs.\ non-contrastive Reinforce---is in Appendix~\ref{app:ablations}.

\begin{table*}[ht]
	\caption{Patch-based repair vs.\ full regeneration in the Solver inner loop, under matched iteration budgets ($N_{\max}=8$). \emph{Solve} is pass@1 (\%); \emph{Iters} is mean Solver iterations per problem (early-exit allowed once all tests pass); \emph{TokSv} is the relative completion-token saving against a fixed reference cost $T_{\text{ref}} = N_{\max}\cdot \bar t_{\text{full}}$, where $\bar t_{\text{full}}$ is the average completion-token count of one full-regeneration draft on that benchmark, i.e.\ the cost of always running the full budget without early-exit and without patch reuse. Both strategies are scored against the same $T_{\text{ref}}$.}
	\label{tab:patch}
	\centering
	\small
	\setlength{\tabcolsep}{4pt}
	\renewcommand{\arraystretch}{1.10}
	\begin{tabular}{l l ccc ccc ccc}
		\toprule
		& & \multicolumn{3}{c}{\textbf{CodeContests (165)}}
		& \multicolumn{3}{c}{\textbf{APPS (1{,}000)}}
		& \multicolumn{3}{c}{\textbf{AetherCode (400)}} \\
		\cmidrule(lr){3-5}\cmidrule(lr){6-8}\cmidrule(lr){9-11}
		Backbone & Strategy & Solve & Iters & TokSv & Solve & Iters & TokSv & Solve & Iters & TokSv \\
		\midrule
		\multirow{2}{*}{GPT-5.4}
		    & \cellcolor{rowtintA}Full regeneration       & \cellcolor{rowtintA}75.76 & \cellcolor{rowtintA}5.18 & \cellcolor{rowtintA}67.4 & \cellcolor{rowtintA}62.10 & \cellcolor{rowtintA}4.41 & \cellcolor{rowtintA}48.6 & \cellcolor{rowtintA}41.25 & \cellcolor{rowtintA}5.32 & \cellcolor{rowtintA}43.7 \\
		    & \cellcolor{macaronlemon}\textbf{Patch repair}   & \cellcolor{macaronlemon}\textbf{82.42} & \cellcolor{macaronlemon}\textbf{3.74} & \cellcolor{macaronlemon}\textbf{91.2} & \cellcolor{macaronlemon}\textbf{67.70} & \cellcolor{macaronlemon}\textbf{3.16} & \cellcolor{macaronlemon}\textbf{88.5} & \cellcolor{macaronlemon}\textbf{49.25} & \cellcolor{macaronlemon}\textbf{3.95} & \cellcolor{macaronlemon}\textbf{86.3} \\
		\midrule
		\multirow{2}{*}{Claude Opus 4.6}
		    & \cellcolor{rowtintA}Full regeneration       & \cellcolor{rowtintA}73.94 & \cellcolor{rowtintA}5.27 & \cellcolor{rowtintA}68.9 & \cellcolor{rowtintA}64.30 & \cellcolor{rowtintA}4.48 & \cellcolor{rowtintA}49.8 & \cellcolor{rowtintA}45.50 & \cellcolor{rowtintA}5.41 & \cellcolor{rowtintA}45.2 \\
		    & \cellcolor{macaronlemon}\textbf{Patch repair}   & \cellcolor{macaronlemon}\textbf{80.61} & \cellcolor{macaronlemon}\textbf{3.86} & \cellcolor{macaronlemon}\textbf{92.0} & \cellcolor{macaronlemon}\textbf{69.30} & \cellcolor{macaronlemon}\textbf{3.24} & \cellcolor{macaronlemon}\textbf{89.4} & \cellcolor{macaronlemon}\textbf{53.75} & \cellcolor{macaronlemon}\textbf{4.03} & \cellcolor{macaronlemon}\textbf{87.5} \\
		\bottomrule
	\end{tabular}
\end{table*}

\subsection{Oracle and Hacker Diagnostic Contribution}
\label{sec:exp:diagnostic}

\begin{figure*}[ht]
	\centering
	\begin{subfigure}[c]{0.48\textwidth}
		\centering
		\scriptsize
		\setlength{\tabcolsep}{4pt}
		\renewcommand{\arraystretch}{1.35}
		\begin{tabular}{llccc}
			\toprule
			Backbone & Config. & Det'd & Pres'd & Str. \\
			\midrule
			\rowcolor{rowtintA}\multirow{3}{*}{Claude Opus 4.6} & Oracle & 82.4 & \textbf{96.8} & 8.5 \\
			\rowcolor{rowtintB} & Hacker & 87.6 & 92.9 & 11.4 \\
			\rowcolor{macaronlemon} & Both & \textbf{92.8} & 96.0 & \textbf{19.6} \\
			\midrule
			\rowcolor{rowtintA}\multirow{3}{*}{GPT-5.4} & Oracle & 79.8 & \textbf{96.1} & 7.4 \\
			\rowcolor{rowtintB} & Hacker & 85.2 & 91.8 & 10.2 \\
			\rowcolor{macaronlemon} & Both & \textbf{90.6} & 95.4 & \textbf{17.9} \\
			\midrule
			\rowcolor{rowtintA}\multirow{3}{*}{DeepSeek V4 Pro} & Oracle & 76.9 & \textbf{95.2} & 6.3 \\
			\rowcolor{rowtintB} & Hacker & 82.7 & 90.5 & 8.8 \\
			\rowcolor{macaronlemon} & Both & \textbf{88.1} & 94.1 & \textbf{15.6} \\
			\bottomrule
		\end{tabular}
		\subcaption{Diagnostic quality of Oracle and Hacker.}
		\label{fig:oracle_hack}
	\end{subfigure}%
	\hspace{0.02\textwidth}%
	\begin{subfigure}[c]{0.50\textwidth}
		\centering
		\includegraphics[width=\linewidth]{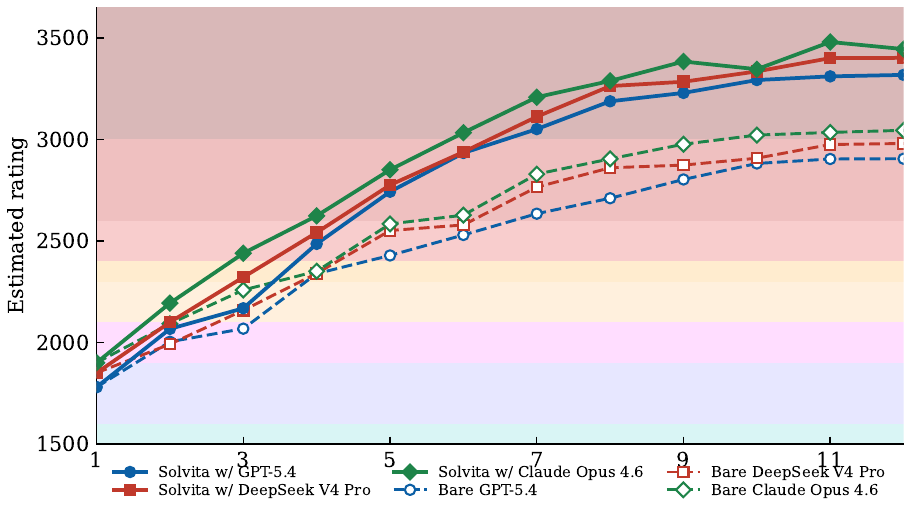}
		\subcaption{Codeforces rating-estimate progression across chronological contest rounds (x-axis).}
		\label{fig:cf}
	\end{subfigure}
	\caption{Oracle/Hacker diagnostics and Codeforces evaluation across three backbones (Claude Opus 4.6, GPT-5.4, DeepSeek V4 Pro). \textbf{(\subref{fig:oracle_hack})} Det'd: wrong solutions detected; Pres'd: correct solutions preserved; Str.: Solvita-rejects/official-accepts disagreement problems confirmed as stronger-test cases. \textbf{(\subref{fig:cf})} Codeforces-style rating-estimate progression for the three Solvita backbones (solid) and the same backbones run bare (dashed), overlaid on the canonical Codeforces tier bands.}
	\label{fig:diag_and_cf}
\end{figure*}

Figure~\ref{fig:oracle_hack} isolates the two diagnostic modules across three backbones on a held-out set with known correctness labels, reporting wrong-solution detection, correct-solution preservation, and confirmed stronger-test rates. The Oracle alone is conservative and preserves correct solutions well, but misses subtle implementation bugs that only adversarial inputs expose. The Hacker alone detects more wrong solutions and surfaces more official-accept disagreements that survive accepted-solution cross-checking and manual validation. Combining the two gives the highest detection and stronger-test confirmation rates for every backbone while maintaining high preservation. The cross-backbone ordering follows the backbone strength in Figure~\ref{fig:cf}, but the Oracle--Hacker complementarity is stable across models. Metric definitions and the stronger-test confirmation protocol are given in Appendix~\ref{app:ablations}.

\subsection{Codeforces Competition Evaluation}
\label{sec:exp:cf}

To complement offline benchmarks, we evaluate Solvita on recent Codeforces rounds (Div.~2 and Div.~1+2). Each contest is attempted in a single uninterrupted session within the official time limit, no corrections allowed after the window closes --- the same constraints as human competitors. We use $K = 12$ post-cutoff contests (rounds 952--963, mixed Div.~2 and Div.~1+2), totalling $76$ problems across A--F slots; contests were selected purely chronologically from the first 12 post-training-cutoff Codeforces rounds with publicly available official editorials, with no per-contest filtering. Figure~\ref{fig:cf} shows the Codeforces-style rating-estimate trajectory for the three backbones (GPT-5.4, DeepSeek V4 Pro, Claude Opus 4.6), computed by inserting each agent into the official standings and inverting the contest-local Elo expectation following CodeElo~\citep{quan2025codeelo} and the classical Elo model~\citep{elo1978rating}. All three Solvita variants converge into the Legendary Grandmaster band ($\geq 3000$) within $K=12$ rounds, while the same three backbones run bare plateau in the high Grandmaster band ($2700$--$2850$), indicating that the gap into the Legendary Grandmaster range comes from the agentic loop rather than the underlying model. The three Solvita curves track each other within $\pm 80$ rating points after round~6 (vs.\ a 140-point spread for the bare backbones), suggesting the gains transfer across backbones rather than depending on a particular model. The exact rank-insertion rule, rating inversion, aggregation, and contest-time configuration are given in Appendix~\ref{app:cf_rating}.

We evaluate Solvita on recent Codeforces rounds (Div.~2 and Div.~1+2). Each contest is attempted in a single uninterrupted session within the official time limit, no corrections allowed after the window closes --- the same constraints as human competitors. Figure~\ref{fig:cf} shows the Codeforces-style rating-estimate trajectory
computed by inserting each agent into the official standings and inverting the contest-local Elo expectation following CodeElo~\citep{quan2025codeelo} and the classical Elo model~\citep{elo1978rating}. All Solvita variants converge into the Legendary Grandmaster band ($\geq 3000$) within roughly a dozen rounds, while the same three backbones run bare plateau in the high Grandmaster band, indicating that the gap into the Legendary Grandmaster range comes from the agentic loop.
The exact rank-insertion rule, rating inversion, aggregation, and contest-time configuration are given in Appendix~\ref{app:cf_rating}.

\section{Related Work}
\label{sec:related}

\textbf{Code generation and self-improving agents.} LLM-based code generation has progressed from single-shot synthesis~\cite{chen2021codex,li2022alphacode} to structured multi-agent pipelines that add planning, retrieval, role separation, hierarchical decomposition, execution-based reranking, repository-level interfaces, and self-debugging or self-repair~\cite{ridnik2024alphacodium,islam2024mapcoder,huang2024agentcoder,le2024codechain,zelikman2023parsel,zhang2023planning,ni2023lever,yang2024sweagent,chen2024selfplay,olausson2024selfrepair}, alongside general-purpose orchestrations and debate~\cite{hong2024metagpt,wu2023autogen,qian2024chatdev,qian2024colearning,liang2024mad}; on top of this, self-improving agents update prompts, rationales, or pipelines through execution feedback or RL~\cite{le2022coderl,zelikman2022star,zelikman2024stop,yao2023tree,zhou2024lats,madaan2023selfrefine,guo2025deepseek,tao2024selfevolution}, but these methods either operate statelessly or improve communication and search rather than a persistent role-aligned memory. Solvita pairs role decomposition with per-agent knowledge networks coupled by adversarial feedback, optimizing a persistent graph by REINFORCE while the LLM stays frozen.

\textbf{Memory and adversarial validation.} Memory-augmented agents store and retrieve past experience through skill libraries, episodic reflection, virtual memory, or graph-structured reasoning~\cite{wang2023voyager,zhao2024expel,shinn2023reflexion,packer2023memgpt,zhang2024memory,lewis2020rag,hu2020hgt,besta2024got}, but flat retrieval and the absence of role specialization remain reported bottlenecks. In parallel, systematic test generation spans fuzzing, equivalence modulo inputs, coverage-guided mutation, LLM-based fuzzing, certified competitive-programming validators, and dedicated hacking pipelines~\cite{le2014compiler,lemieux2023codamosa,deng2023titanfuzz,chen2023codet,liu2023evalplus,schafer2024testpilot,wang2025codecontests,shi2026codehacker}. 
Solvita differs in partitioning experience by agent role with learned edge weights, and in embedding certified test construction and adversarial attacks inside one closed-loop framework where each discovery updates all four knowledge networks at once.

\section{Conclusion}
\label{sec:conclusion}
We introduced Solvita, a framework that overcomes the limitations of stateless code generation by enabling continuous, experience-driven learning for frozen LLMs. By coupling four specialized agents (Planner, Solver, Oracle, Hacker) with dynamic, graph-structured knowledge networks, Solvita translates execution verdicts and adversarial testing into REINFORCE updates, allowing the system to accumulate algorithmic intuition, strategy routing, and debugging experience over time. Empirically, Solvita achieves a new state of the art across rigorous competitive-programming benchmarks---including live Codeforces rounds---nearly doubling the accuracy of single-pass baselines.

\paragraph{Limitations.} Three trade-offs are visible in our runs. (i)~Cold-start cost: the agentic loop is meaningfully more expensive than direct generation per problem, and the knowledge networks need on the order of 5{,}000 training problems before the per-problem cost is amortized into accuracy gains. (ii)~Hacker scope: anti-hash and lattice-based attacks are bounded by the backbone's reasoning horizon, so heavily math-flavored failure modes (number-theoretic invariants, geometric tolerance bugs) remain under-covered. (iii)~Patch-repair drift: on globally flawed candidates the Solver can mislabel a systemic flaw as localized and accumulate inconsistent edits before the iteration budget exhausts; the regression-rate signal in Section~\ref{sec:exp:patch} catches this only post hoc.

\paragraph{Future work.} Three directions follow naturally. First, warm-starting the knowledge networks from open-source experience corpora (editorials, accepted submissions, debugging traces) should shrink the cold-start window. Second, transferring the four-agent decomposition to other verifiable reasoning domains---formal theorem proving, where the Oracle becomes a proof checker and the Hacker searches for counter-models; mathematical olympiad problems, where certified test cases are replaced by symbolic verification; and scientific reasoning with executable simulators---is a direct port of the same closed-loop interface to settings that share competitive programming's verifier-grounded reward structure. Third, the per-step adversarial signal produced by the Hacker is a candidate fine-tuning signal beyond prompt-level updates: investigating whether REINFORCE on knowledge-network weights can be lifted into model-weight updates without losing the role-aligned credit assignment is, to us, the most interesting open question.

\bibliographystyle{unsrtnat}
\bibliography{references}

\appendix
\section*{Appendix}

\section{Data Pipeline Configuration}
\label{app:data:pipeline}

This appendix lists the exact configuration of every step of the filtering pipeline in Section~\ref{sec:data:filtering}.

\paragraph{Per platform difficulty mapping.}
Each source platform exposes difficulty in a different unit; we normalize them into a single ordinal scale used by the difficulty signal in step (1) and the per tag floor in step (3). The mapping covers Codeforces rating bands, AtCoder score and color tier, LeetCode Easy/Medium/Hard, and the CodeContests difficulty tag. \begin{table}[htbp]
\centering
\caption{Per-platform difficulty normalization onto the Codeforces rating scale (800--3500). The same normalized tier is used across all platforms in step (1) and as the $d_{\min}$ cutoff in step (3). LeetCode labels map to a numeric range because they are coarse-grained; AtCoder scores and Codeforces ratings are used directly.}
\label{tab:difficulty-mapping}
\renewcommand{\arraystretch}{1.15}
\begin{tabular}{@{}llll@{}}
\toprule
\textbf{Platform} & \textbf{Native Difficulty Unit} & \textbf{Normalized Tier (CF Rating)} & \textbf{Typical $d_{\min}$ Cutoff} \\
\midrule
Codeforces     & Problem rating (800--3500)          & Same as native                 & 800 / 900 per tag \\
AtCoder        & ABC-A / ABC-B                      & 800                            & 800 \\
               & ABC-C                              & 900--1100                      & 900 \\
               & ABC-D                              & 1200--1300                     & 1200 \\
               & ABC-E                              & 1400--1600                     & 1400 \\
               & ABC-F                              & 1700--1900                     & 1700 \\
               & ABC-G / ARC / AGC                  & 1900+                          & 1900 \\
LeetCode       & Easy                               & 800--900                       & 800 \\
               & Medium                             & 1000--1600                     & 1000 \\
               & Hard                               & 1500+                          & 1400 \\
\bottomrule
\end{tabular}
\vspace{4pt}
\begin{minipage}{0.95\textwidth}
\footnotesize
\textit{Notes.} (1)~Codeforces ratings are on an 800--3500 integer scale derived from the Elo-based rating system. (2)~AtCoder ABC mapping follows community-consensus ranges. (3)~LeetCode mapping follows the widely cited rule that LeetCode difficulty roughly corresponds to Codeforces rating minus 600--700, with community-ratified bands. (4)~The \textbf{Typical $d_{\min}$ Cutoff} column lists the per-tag difficulty floors commonly applied in step (3); the actual value is determined per tag at the 5\textsuperscript{th} percentile of its post-deduplication difficulty distribution.
\end{minipage}
\end{table}

\paragraph{Tag load balancing.}
Step (2) caps each tag at $C_{\max}=2300$ surviving problems. Tags above the cap are subsampled uniformly at random within their difficulty distribution so that the per tag count after balancing stays within a constant factor of the smallest surviving tag.
\begin{table}[htbp]
\centering
\setlength{\tabcolsep}{6pt}
\caption{Per‑tag problem counts before and after tag load balancing (step 2). The fraction kept is computed as count after balancing divided by count before balancing.}
\label{tab:tag-balancing}
\begin{tabular}{@{}lccc@{}}
\toprule
\textbf{Tag} & \textbf{Before Balancing} & \textbf{After Balancing} & \textbf{Fraction Kept} \\
\midrule
implementation               & 3819 & 2300 & 60.2\% \\
math                         & 3200 & 2300 & 71.9\% \\
greedy                       & 2840 & 2300 & 81.0\% \\
dp                           & 2625 & 2300 & 87.6\% \\
data\_structures             & 2329 & 2118 & 90.9\% \\
constructive\_algorithms     & 1759 & 1594 & 90.6\% \\
brute\_force                 & 1429 & 1302 & 91.1\% \\
graphs                       & 1416 & 1359 & 96.0\% \\
sortings                     & 1409 & 1290 & 91.6\% \\
binary\_search               & 1403 & 1336 & 95.2\% \\
dfs\_and\_similar            & 1357 & 1168 & 86.1\% \\
trees                        & 1225 & 1155 & 94.3\% \\
strings                      & 1154 & 1007 & 87.2\% \\
number\_theory               & 994 & 943 & 94.8\% \\
combinatorics                & 804 & 788 & 98.0\% \\
\bottomrule
\end{tabular}
\vspace{4pt}
\begin{minipage}{0.95\textwidth}
\footnotesize
\textit{Notes.} (1)~Counts after balancing are computed after applying the per‑tag cap $C_{\max}=2300$. (2)~Fractions are computed per tag as (after balancing) / (before balancing). (3)~Statistics are computed over all tag occurrences; a problem carrying multiple tags contributes to each of its tags. The table lists the 15 most frequent tags in the corpus.
\end{minipage}
\end{table}

\paragraph{Embedding and deduplication threshold.}
Step (3) uses \texttt{text-embedding-3-large} with cosine similarity computed inside each tag bucket. The duplicate threshold is $\delta=0.93$, chosen to maximize precision on a manually labeled validation set of 500 candidate pairs.

\paragraph{Retained tags and per tag difficulty floors.}
After steps (1)--(2) we retain $T=107$ algorithmic tags. Each tag $\tau$ has its own difficulty floor $d_{\min}^{(\tau)}$ used in step (4), set so that the easiest surviving instance of $\tau$ is still non-trivial for the base LLM.
\begin{table}[htbp]
\centering
\caption{Per-tag statistics and difficulty floors.}
\label{tab:tag-floors}
\renewcommand{\arraystretch}{1.1}
\begin{tabular}{@{}lccc@{}}
\toprule
\textbf{Tag} & \textbf{Retained after Step (3)} & \textbf{$d_{\min}^{(\tau)}$} & \textbf{Retained after Step (4)} \\
\midrule
implementation               & 1817 & 1400 & 890 \\
math                         & 1886 & 1600  & 719 \\
greedy                       & 1925 & 1600  & 951 \\
dp                           & 1891 & 1600 & 1503 \\
data\_structures             & 1783 & 1600 & 1366 \\
constructive\_algorithms     & 1322 & 1600  & 1109 \\
brute\_force                 & 946 & 1400  & 782 \\
graphs                       & 974  & 1600 & 640 \\
sortings                     & 905  & 1400  & 653 \\
binary\_search               & 1003  & 1600  & 727 \\
dfs\_and\_similar            & 957  & 1600 & 681 \\
trees                        & 806  & 1600 & 499 \\
strings                      & 745  & 1300  & 453 \\
number\_theory               & 684  & 1400 & 394 \\
combinatorics                & 526  & 1400 & 375 \\
\bottomrule
\end{tabular}
\vspace{4pt}
\begin{minipage}{0.95\textwidth}
\footnotesize
\textit{Notes.}(1)~Statistics are computed over all tag occurrences; a problem carrying multiple tags contributes to each of its tags. The table lists the 15 most frequent tags in the corpus.
\end{minipage}
\end{table}

\paragraph{Per platform raw and surviving counts.}
\begin{table}[htbp]
\centering
\footnotesize
\setlength{\tabcolsep}{3pt}
\caption{Per‑platform problem counts at each filtering stage. The total row sums to the corpus‑wide figures reported in Section~\ref{sec:data:filtering}.}
\label{tab:per-platform-counts}
\begin{tabular}{@{}lccccc@{}}
\toprule
\textbf{Platform} & \textbf{Raw Collected} & \textbf{After Completeness} & \textbf{After Balance} & \textbf{After Deduplication} & \textbf{After Difficulty Pruning} \\
\midrule
Codeforces     & 15862 & 15573 & 11634 & 8841 & 6325 \\
AtCoder        & 4187 & 3984 & 3471 & 3316 & 756 \\
Aizu           & 2645 & 2524 & 2276 & 2242 & 249 \\
Others         & 7324 & 2631 & 2105 & 2104 & 687 \\
\midrule
\textbf{Total} & 30,018 & 24,712 & 19,486 & 16,503 & 8,017 \\
\bottomrule
\end{tabular}
\vspace{4pt}
\begin{minipage}{0.95\textwidth}
\footnotesize
\textit{Notes.} (1)~The total row reproduces the corpus‑level figures from the main text. (2)~The final column reflects the corpus after applying the per‑tag difficulty floors $d_{\min}^{(\tau)}$ (Table~\ref{tab:tag-floors}); a problem may be counted against its primary platform even if it was retained under different tags post‑balancing.
\end{minipage}
\end{table}

\section{Contextual Bandit Policy Details}
\label{app:bandit}

Each agent's knowledge network uses the contextual bandit policy described in Section~\ref{sec:method:arch}. The learning rate is $\alpha=0.01$, rewards lie in $r \in [-1, 1]$, and a tag-overlap bonus of $+0.05$ per matching tag provides a prior. Feature keys encode the agent's current finite-state-machine position (e.g., \texttt{FSM:SOLVE\_DRAFT}), any failure type from the previous iteration (\texttt{FAIL:TIMEOUT}), and problem-level tags (\texttt{TAG:dp}). Parameters are persisted in JSON with atomic file-locking writes. Items whose running average reward drops below $-0.3$ after 20 or more uses are automatically deprecated.

\section{Oracle Reward: Failure-Path Details}
\label{app:oracle}

A problem instance is $x = (d, c, p, \kappa)$ where $d$ is the description, $c$ the constraints, $p$ the public samples, and $\kappa$ the structural context. The Oracle produces a supervision artifact $y = (F, f^*, T, V, A, m)$ containing the candidate family set, selected family, certified tests, verifier provenance, acceptance indicator, and metadata. Family selection uses the bandit policy of Section~\ref{sec:method:arch},
\begin{equation}
	f^{*}(x) \;=\; \arg\max_{f \,\in\, F(x)}\; s_f(x),
	\qquad
	s_f(x) \;=\; b_f \,+\, \sum_{k \,\in\, \Phi(x)} W_{k,\,f},
	\label{eq:oracle_family}
\end{equation}
where $\Phi(x)$ is the set of active feature keys, and the acceptance gate follows Eq.~\ref{eq:oracle_gate}. The successful-artifact reward defined in Section~\ref{sec:method:oracle} is augmented with the following explicit negative penalties on failure paths, which are needed to keep the bandit signal informative when no certified test survives:
\begin{equation}
	r_{\text{oracle}}^{\text{fail}} \;=\;
	\begin{cases}
		-1.0, & N_{\text{cert}} = 0 \;(\text{crash or severe error}),\\[2pt]
		-0.7, & N_{\text{cert}} = 0 \;(\text{self-check failure}),\\[2pt]
		-0.6, & \text{no valid tests or state not ready}.
	\end{cases}
	\label{eq:oracle_reward_fail}
\end{equation}
For full certification ($\rho = 1$), the bonus $r_{\text{verify}} \in \{+1.0, -0.2, -0.5\}$ depends on whether the independent judge agrees, partially agrees, or contradicts the certified suite.

\section{Hacker Reward: Degenerate-Round Details}
\label{app:hack}

Let $V$ be all sandbox verdicts, $V_{\text{valid}}$ the valid-input subset, $V_{\text{break}} \subseteq V_{\text{valid}}$ those exposing a failure, and $c$ the compilation failure count. The default per-round reward is the clipped composition of Eq.~\ref{eq:hack_reward}; on degenerate rounds (\texttt{Gen\_Failed} with no valid verdict) the explicit correction $r = -0.6 - \min(0.3,\, 0.1\,c)$ replaces the default so that repeated generator failures still produce a usable gradient. The judge is resolved in the same priority order as the Oracle: custom checker, correct-solution runner, exact match.

\paragraph{Component weights.} The Hacker reward of Eq.~\ref{eq:hack_reward} is a weighted combination of the valid-input rate $g_{\text{valid}}$, the break rate $g_{\text{break}}$, and the average severity $g_{\text{sev}}$, minus a compile-failure penalty $\kappa(c)$. The default weights and penalty are tabulated in Table~\ref{tab:hack_weights}. The bulk of the budget sits on $w_b$ so that actually breaking the candidate dominates the gradient, while $w_v$ keeps the router from collapsing to high-severity but invalid-by-validator inputs.

\begin{table}[h]
  \centering
  \small
  \begin{tabular}{lcc}
    \toprule
    Quantity & Symbol & Value \\
    \midrule
    Valid-input weight     & $w_v$         & $0.20$ \\
    Break-rate weight      & $w_b$         & $0.55$ \\
    Severity weight        & $w_s$         & $0.25$ \\
    Compile penalty (per failure, capped at $0.3$) & $\kappa(c)$ & $\min(0.3,\,0.1\,c)$ \\
    Reward clip range      & ---           & $[-1.0,\,+1.0]$ \\
    \bottomrule
  \end{tabular}
  \caption{Hacker reward component weights and compile-failure penalty schedule.}
  \label{tab:hack_weights}
\end{table}

\paragraph{Per-verdict severity.} The severity table $\omega(\cdot)$ maps each per-input verdict to a scalar in $[0,1]$ that feeds the round-level $g_{\text{sev}} = \frac{1}{|V_{\text{break}}|}\sum_{v \in V_{\text{break}}} \omega(v)$. Table~\ref{tab:hack_severity} lists the values used across all reported runs; they encode the rough cost-to-fix ordering used by competitive-programming judges (a wrong-answer is cheap to triage, a sandbox crash is expensive).

\begin{table}[h]
  \centering
  \small
  \begin{tabular}{lc}
    \toprule
    Verdict & $\omega(\cdot)$ \\
    \midrule
    Wrong Answer (WA)            & $0.50$ \\
    Time Limit Exceeded (TLE)    & $0.65$ \\
    Memory Limit Exceeded (MLE)  & $0.75$ \\
    Runtime Error (RE)           & $0.85$ \\
    Sandbox crash / fatal signal & $1.00$ \\
    \bottomrule
  \end{tabular}
  \caption{Per-verdict severity weights used by the Hacker reward. Values were calibrated once on a 200-problem dev split and held fixed across all reported runs.}
  \label{tab:hack_severity}
\end{table}

\paragraph{Degenerate-round sanity.} When $|V_{\text{valid}}| = 0$ the round contains no informative break signal, but the generator may still have learned (or unlearned) something --- e.g.\ producing inputs the validator rejects en masse. The fixed $-0.6$ baseline plus the linear compile penalty ensures the bandit signal stays bounded and nonzero so the route weights still update; without this correction, repeated \texttt{Gen\_Failed} rounds would silently zero out the gradient and freeze the router on its current arm.

\section{Prompt Details}
\label{app:prompts}

This appendix collects the actual prompt templates used by every agent in our experiments. All prompts are stored in a single YAML file (\texttt{config/prompt\_template.yaml}) and rendered through a placeholder substitution layer (\texttt{<KEY>} tokens are filled at call time). For brevity we show the role-defining portions; cross-cutting boilerplate (output schemas, fast-I/O reminders, JSON-escaping warnings) is preserved verbatim where it materially affects behavior. The same prompts are used across all five backbones in our experiments.

\paragraph{Conventions.} Unless noted otherwise, every JSON-output prompt below requires \emph{strict} JSON: no Markdown fences, no commentary outside the object, and embedded newlines/tabs/backslashes escaped (\texttt{\textbackslash n}, \texttt{\textbackslash t}, \texttt{\textbackslash\textbackslash}). C++ outputs follow a single template: \texttt{\#include}-headers only (no \texttt{<bits/stdc++.h>}), C++17, fast I/O, with every major data-structure allocation justified against the input bounds; we explicitly highlight these only where the prompt diverges from this default. The TLE budget rule used throughout is the standard $10^{8}$~ops/sec heuristic: \texttt{iterations / 10\textasciicircum{}8 $\le$ time\_limit\_seconds}.

\subsection{Planner}

The Planner first reduces the raw problem statement to a canonical form and a tag set. Both the system message and the user message are short and JSON-bound so that downstream agents can parse the output deterministically.

\begin{promptbox}{Planner --- abstract\_problem (system)}
You extract a precise canonical formulation and algorithmic tags for a competitive programming task.\\
Respond with a single JSON object only. No markdown fences.
\end{promptbox}

\begin{promptbox}{Planner --- abstract\_problem (user)}
Original problem:\\
\textless PROBLEM\_DESC\textgreater\\
\\
Known problem types (hints): \textless PROBLEM\_TYPES\textgreater\\
\\
Constraints JSON:\\
\textless CONSTRAINTS\_JSON\textgreater\\
\\
Primary tags (level-1, coarse topics -- prefer at least one when it fits):\\
\textless TAG\_WHITELIST\_LEVEL1\textgreater\\
\\
Fine-grained tags (level-2 -- optional; add when they precisely match the task):\\
\textless TAG\_WHITELIST\_LEVEL2\textgreater\\
\\
Allowed tags reference (union of level-1 and level-2 vocabularies):\\
\textless TAG\_WHITELIST\textgreater\\
\\
Return JSON with this exact shape:\\
\{\\
\hspace*{1em}"canonical\_problem": \{\\
\hspace*{2em}"objective": "string",\\
\hspace*{2em}"inputs": \{\},\\
\hspace*{2em}"outputs": \{\},\\
\hspace*{2em}"constraints": \{"normalized": \{\}, "derived": []\},\\
\hspace*{2em}"required\_properties": [],\\
\hspace*{2em}"edge\_cases": []\\
\hspace*{1em}\},\\
\hspace*{1em}"algorithmic\_tags\_level1": ["primary tags from the level-1 list only"],\\
\hspace*{1em}"algorithmic\_tags\_level2": ["optional fine-grained tags from the level-2 list only"],\\
\hspace*{1em}"abstract\_confidence": 0.0,\\
\hspace*{1em}"abstract\_trace": \{"rationale": "why this abstraction fits", "warnings": []\}\\
\}\\
Rules:\\
- canonical\_problem must be self-contained and precise.\\
- algorithmic\_tags\_level1 are the main coarse topic labels for this task; use only entries from the level-1 list above.\\
- algorithmic\_tags\_level2 are optional finer-grained labels; use only entries from the level-2 list above, and do not restate the same idea already covered by level-1.\\
- If a tag could fit both levels, put it in level-1 only and omit it from level-2.\\
- abstract\_confidence in [0,1] reflects how well the statement constraints are satisfied.\\
- abstract\_trace must briefly explain tagging and any ambiguity.
\end{promptbox}

\subsection{Solver: skill selection}

After the canonical form is available, the Solver consults its three-layer QMS knowledge network (Section~\ref{sec:method:solver}) and asks the LLM to pick a small number of skill identifiers and emit a sub-problem DAG. The strict JSON contract (no fences, exact id copying, no commentary) is enforced because the result is used as a direct lookup key into the skill table.

\begin{promptbox}{Solver --- solver\_skill\_selection (system)}
You output exactly one JSON object and nothing else.\\
Required keys: (1) "selected\_skill\_ids" (array of strings) --- copy ids exactly from the candidate list;\\
(2) "subproblem\_dag" (object) with "nodes" (array of \{id, description\}) and "edges" (array of \{from\_id, to\_id\}).\\
The DAG must be acyclic: edges mean dependency (from\_id before to\_id).\\
Use abstract algorithmic descriptions for nodes (not story fluff).\\
If no candidate skill applies, use [] for selected\_skill\_ids; you may still output a useful\\
subproblem\_dag for the \textbf{current} problem (abstract nodes/edges), or use empty arrays if no decomposition helps.\\
No markdown fences, no commentary, no text before or after the JSON.
\end{promptbox}

\begin{promptbox}{Solver --- solver\_skill\_selection (user)}
You are selecting algorithmic \textbf{skills} (reference snippets) for a competitive programming task, and you must\\
also output a \textbf{sub-problem DAG} that decomposes the \textbf{current} problem into abstract sub-goals.\\
Use the \textbf{problem summary}, the \textbf{activated graph context} (similar Q / M evidence), and the \textbf{candidate list}\\
($\rho$ ranks retrieval relevance).\\
Choose \textbf{between \textless MIN\_SELECT\textgreater\ and \textless MAX\_SELECT\textgreater\ (inclusive)} skills that genuinely help solve the current task.\\
If \textbf{no} candidate is relevant, return an empty list for skills and empty DAG nodes/edges. Otherwise prefer \textbf{at least}\\
\textless MIN\_SELECT\textgreater\ skills when that many good fits exist in the list.\\
\\
\#\# Problem summary\\
\\
\textless PROBLEM\_SUMMARY\textgreater\\
\\
\#\# Activated graph context (top-Q similarity and linked M-nodes)\\
Structured evidence from the skill graph: similar problems (Q) and their logic pipelines (M).\\
Align skill choices with this retrieval context.\\
\\
\textless ACTIVATED\_GRAPH\_CONTEXT\textgreater\\
\\
\#\# Candidate skills (id, title, description, $\rho$)\\
\\
\textless CANDIDATE\_SKILLS\_BLOCK\textgreater\\
\\
\#\# Valid skill ids (copy exactly, one per line)\\
Your JSON strings must match one of these lines \textbf{verbatim} (including hyphens):\\
\\
\textless VALID\_SKILL\_IDS\_BLOCK\textgreater\\
\\
\#\# Output format (strict)\\
Return \textbf{only} one JSON object and \textbf{nothing else} --- no markdown fences, no commentary, no reasoning.\\
Required keys:\\
\hspace*{1em}- "selected\_skill\_ids": string[] --- ids copied verbatim from the candidate list.\\
\hspace*{1em}- "subproblem\_dag": \{ "nodes": [...], "edges": [...] \}\\
\hspace*{2em}- nodes: each item is \{ "id": string (e.g. P1), "description": string \} --- abstract algorithmic sub-goal, not story text.\\
\hspace*{2em}- edges: each item is \{ "from\_id": string, "to\_id": string \} --- dependency (from\_id must be solved before to\_id). The graph must be a DAG (no cycles).\\
If selected\_skill\_ids is [], you may still output a meaningful subproblem\_dag for this task,\\
or use empty nodes/edges only when no decomposition is useful.
\end{promptbox}

\begin{promptbox}{Solver --- solver\_skill\_selection (user, minimal retry)}
Output \textbf{only} one JSON object. No markdown fences, no reasoning text.\\
Include "selected\_skill\_ids" and "subproblem\_dag" \{ "nodes": [...], "edges": [...] \} (DAG, acyclic).\\
Choose \textless MIN\_SELECT\textgreater--\textless MAX\_SELECT\textgreater\ ids from the list below (or [] if none apply).\\
\\
\#\# Task (short)\\
\\
\textless PROBLEM\_SUMMARY\_SHORT\textgreater\\
\\
\#\# Candidates (copy ids exactly from the bottom list)\\
\\
\textless CANDIDATE\_LINES\_MINIMAL\textgreater\\
\\
\#\# Valid skill ids (verbatim, one per line)\\
\\
\textless VALID\_SKILL\_IDS\_BLOCK\textgreater\\
\\
Example:\\
\{\\
\hspace*{1em}"selected\_skill\_ids":["xxxxxxxx-xxxx-xxxx-xxxx-xxxxxxxxxxxx"],\\
\hspace*{1em}"subproblem\_dag":\{\\
\hspace*{2em}"nodes":[\{"id":"P1","description":"Parse input and build data structure"\}],\\
\hspace*{2em}"edges":[]\\
\hspace*{1em}\}\\
\}
\end{promptbox}

\subsection{Solver: code generation and patch repair}

The Solver alternates between full-program generation (initial draft) and search-and-replace patch repair (subsequent iterations, up to $N_{\max}=8$). A separate \texttt{patch\_decision} prompt routes between the two modes based on whether the failure looks localized or systemic. The generation prompt embeds an explicit \emph{resource audit} requirement so that the model rejects sketches that would be infeasible at the stated constraint bounds.

\needspace{20\baselineskip}
\begin{promptboxbreak}{Solver --- generate\_code.initial}
You are solving a competitive programming problem. Apply the following three skills in order before writing any code.\\
\\
\#\# Problem Context\\
\\
Problem:\\
\textless PROBLEM\_DESC\textgreater\\
\\
\textless ABSTRACT\_TAGS\_LEVEL2\_BLOCK\textgreater\\
\\
Proposed algorithm:\\
\textless ALGORITHM\textgreater\\
\\
Implementation steps:\\
\textless STEPS\textgreater\\
\\
\textless CONSTRAINTS\_BLOCK\textgreater\\
\\
\textless PUBLIC\_BLOCK\textgreater\\
\\
\textless GEN\_BLOCK\textgreater\\
\\
\textless SOLVER\_GRAPH\_BLOCK\textgreater\\
\\
\textless SELF\_VALIDATION\_BLOCK\textgreater\\
\\
\textless MEMORY\_ADVICE\textgreater\\
\\
---\\
\\
\#\# Skill 1: Brainstorming --- Design Before Implementation\\
\\
\textless HARD-GATE\textgreater\\
Do NOT write any C++ code until you have completed the design steps below.\\
This applies regardless of how simple the problem appears.\\
\textless /HARD-GATE\textgreater\\
\\
Produce a \textbf{concise} algorithm design (target: under 400 words total) covering:\\
\\
1. \textbf{Core approach}: In 1--3 sentences, name the EXACT technique and data structures. Be specific: not "tree DP" but "rerooting DP storing subtree expected values"; not "greedy" but "leftist-heap merge tracking incremental cost per frontier". If you cannot name the specific data structure and key operation in one sentence, you have not converged --- try a different angle before proceeding.\\
2. \textbf{Complexity + TLE check}: State time O(?) and space O(?). Estimate operations concretely. Apply the \textbf{10\textasciicircum{}8 simple ops/sec rule for C++}: if `operations / 10\textasciicircum{}8 > time\_limit\_seconds`, the approach is TLE --- redesign. Do not rationalize that constants will save you.\\
3. \textbf{Correctness invariant}: The one-sentence key insight that makes this approach correct.\\
\\
If the proposed algorithm in "Implementation steps" above would lead to an unsafe or hard-to-implement solution, you are free and encouraged to choose a better approach instead.\\
\\
---\\
\\
\#\# Skill 2: Test-Driven Development --- Verify Algorithm Before Coding (Red Phase)\\
\\
\textbf{Iron Law: NO CODE WITHOUT A VERIFIED ALGORITHM.}\\
\\
Trace your chosen algorithm on each public sample input:\\
- Show key intermediate values and the final output your algorithm produces.\\
- Confirm it matches the expected output.\\
\\
If any sample trace fails: \textbf{STOP. Redesign the algorithm. Do not write code for a failing approach.}\\
\\
Only proceed to implementation after all samples pass the mental trace.\\
\\
---\\
\\
\#\# Skill 3: Writing Plans --- Implementation Plan Before Coding\\
\\
Before writing any code, decompose the implementation into bite-sized steps. For this problem, list:\\
\\
1. \textbf{Data structures to define} (e.g., struct Node, priority\_queue type, etc.)\\
2. \textbf{Main function steps} in order (e.g., "read input -> build trie -> run DP -> output")\\
3. \textbf{Any tricky sub-routine} that needs careful implementation (e.g., "merge two heaps", "BFS with state compression")\\
\\
This plan is a 5--10 line outline, not prose. Write it down before touching code.\\
\\
---\\
\\
\#\# Skill 4: Implementation (Green Phase) + Verification Before Completion\\
\\
Now implement following your plan. Requirements:\\
- Use standard C++ (C++17)\\
- Do NOT use non-standard headers like \#include \textless bits/stdc++.h\textgreater\\
- Include all necessary standard headers explicitly\\
- Implement fast I/O\\
- Every major data structure allocation must be defensible given the maximum input bounds --- state its size explicitly in a comment if non-obvious\\
- Prefer streaming, rolling-state, or sparse representations when full dense tables would exceed memory\\
- If the "Implementation steps" sketch above would lead to an unsafe implementation, adapt --- do not follow it literally\\
\\
After writing the code, before outputting it, verify:\\
- \textbf{Time (strict)}: Count the dominant loop's iterations for maximum input. Use the \textbf{10\textasciicircum{}8 ops/sec rule}: `iterations / 10\textasciicircum{}8 <= time\_limit\_seconds`. If you wrote O(N * Q) and N=Q=10\textasciicircum{}5, that is 10\textasciicircum{}10 ops = 100s --- TLE. Redesign if this check fails. Do NOT assume constants will save you.\\
- \textbf{Memory}: For each major allocation, is the total within the memory limit?\\
- \textbf{Edge cases}: Does the code handle n=1 and n=max?\\
\\
If any check fails: fix the code first.\\
\\
---\\
\\
Output your full response in this exact structure:\\
\\
\#\#\# Design\\
\texttt{[}Algorithm design, sample traces, and implementation plan from Skills 1--3 above\texttt{]}\\
\\
\#\#\# Solution\\
```cpp\\
\texttt{[}Complete C++17 solution\texttt{]}\\
```
\end{promptboxbreak}

\needspace{20\baselineskip}
\begin{promptboxbreak}{Solver --- generate\_code.think}
You are analyzing a competitive programming problem. Your task is ONLY to determine the best algorithm --- do NOT write any C++ code in this response.\\
\\
\#\# Discipline (non-negotiable)\\
\\
\textbf{TDD-style rule for competitive programming}: Before declaring an algorithm correct, you must demonstrate it on actual computation, not mental simulation. The Iron Law:\\
\\
\textgreater\ NO `VERDICT: PROCEED` WITHOUT A `\textless run\_cpp\textgreater` OR `\textless run\_python\textgreater` BLOCK THAT EXECUTED AND PRODUCED THE EXPECTED OUTPUT IN THIS TURN.\\
\\
"Mental traces" are forbidden as the sole evidence. If your reasoning concludes "and clearly the answer is X", you must run code that prints X.\\
\\
\textbf{Brainstorming rule (Codex `brainstorming` skill)}: For any non-trivial problem, list 2--3 candidate algorithms with their complexity classes BEFORE picking one. State the trade-off (time vs. space vs. implementation difficulty) and pick the one that strictly fits the constraints. If only one approach fits, say so explicitly.\\
\\
\textbf{Verification rule (Codex `verification-before-completion` skill)}: Evidence before claims. If you haven't run a verification command in this message, you cannot claim it passes. "It should work" is not evidence --- `stdout matches expected on tests T1, T2, T3` is evidence.\\
\\
\textbf{Anti-pattern: confidently wrong math.} A formula that "looks right" can hide an algebraic step that breaks under specific inputs. The cure is brute force comparison on small random inputs, not more squinting at the formula. Two-parameter parameterizations of n-parameter structures are a classic trap; if your derivation reduced parameter count, write a brute force enumerator and confirm the count agrees on n=2, n=3, n=4 before trusting your reduction.\\
\\
\textbf{Anti-pattern: TLE that "should fit".} A complexity that "should run in 0.5s" often runs in 8s due to constant factors (cache misses, branch mispredictions, integer division). When in doubt, compile a draft via `\textless run\_cpp\textgreater` on a worst-case-shaped input near the constraint bound. If it doesn't finish in 5s in the sandbox, it won't finish on the judge.\\
\\
---\\
\\
\#\# Problem Context\\
\\
Problem:\\
\textless PROBLEM\_DESC\textgreater\\
\\
\textless ABSTRACT\_TAGS\_LEVEL2\_BLOCK\textgreater\\
\\
Proposed algorithm hint (may be incomplete or wrong --- use your own judgment):\\
\textless ALGORITHM\textgreater\\
\\
\textless CONSTRAINTS\_BLOCK\textgreater\\
\\
\textless PUBLIC\_BLOCK\textgreater\\
\\
\textless ORACLE\_STATUS\_BLOCK\textgreater\\
\\
\textless HARD\_GATE\_BLOCK\textgreater\\
\\
\textless MEMORY\_ADVICE\textgreater\\
\\
---\\
\\
\#\# Your Task: Algorithm Convergence (NO CODE)\\
\\
You can verify your reasoning by embedding tool blocks in your reply. The harness executes them and feeds the output back to you:\\
\\
- `\textless run\_python\textgreater...\textless /run\_python\textgreater` --- Python sandbox (`sys`, `math`, `itertools`, `collections`, `bisect`, `heapq`, `random`); 5s CPU. Use it to write a tiny brute-force solver and check small cases.\\
- `\textless run\_cpp\textgreater...\textless /run\_cpp\textgreater` --- Real C++ compilation + execution (full standard library; the same `g++ -O2 -std=c++17` compiler the judge uses; 5s CPU). To pipe stdin, wrap it as:\\
\hspace*{1em}```\\
\hspace*{1em}\textless run\_cpp\textgreater\\
\hspace*{1em}INPUT\_BEGIN\\
\hspace*{1em}3\\
\hspace*{1em}1 2 3\\
\hspace*{1em}INPUT\_END\\
\hspace*{1em}\#include \textless bits/stdc++.h\textgreater\\
\hspace*{1em}using namespace std;\\
\hspace*{1em}int main()\{ ... \}\\
\hspace*{1em}\textless /run\_cpp\textgreater\\
\hspace*{1em}```\\
\hspace*{1em}The C++ tool is the BEST way to detect TLE early: pick a worst-case-shaped input near the constraint bound and run a candidate implementation. If it doesn't finish in 5s in the sandbox, it won't finish on the judge.\\
\\
\textbf{Strongly recommended workflow}: write a Python brute force (small inputs, exponential is OK), write a candidate C++ implementation, run BOTH on the same small random inputs in tool blocks, and only declare PROCEED when their outputs agree on at least 5 random tests AND your C++ runs within budget on a worst-case-shaped input.\\
\\
Think through the problem and converge on the best algorithm. Produce:\\
\\
1. \textbf{Algorithm} (1--3 sentences, maximally specific): Name the exact data structure and core operation --- e.g., "leftist-heap merge tracking incremental cost per trie frontier node", not just "greedy" or "DP". If you are not certain, name the best candidate and explain why it fits.\\
\\
2. \textbf{Complexity + TLE check}: State O(time) and O(space). Apply the 10\textasciicircum{}8 ops/sec rule: `ops / 10\textasciicircum{}8 <= time\_limit\_seconds`. If this check fails, redesign before continuing.\\
\\
3. \textbf{Correctness invariant}: One sentence stating why this approach produces the correct answer.\\
\\
4. \textbf{Sample trace}: Trace your algorithm on the first public sample. Show key intermediate values and confirm the output matches.\\
\\
5. \textbf{Implementation sketch} (3--5 bullets): The key data structures to define and the main function's control flow --- specific enough that no further design decisions are needed when writing code.\\
\\
6. \textbf{Verdict} --- end your response with EXACTLY ONE of these two lines, on its own line:\\
\hspace*{1em}- `VERDICT: PROCEED` --- if your algorithm clearly satisfies the time/memory limits and your derivation is sound.\\
\hspace*{1em}- `VERDICT: REDESIGN\_NEEDED --- \textless one-sentence reason\textgreater` --- if your complexity analysis exceeds the time limit, your formula has a step you cannot verify, or you discovered an inconsistency. Be honest: choosing REDESIGN\_NEEDED is preferred over forcing a flawed solution forward.\\
\\
Stop here. Do NOT output any C++ code. The next step will ask for the implementation.
\end{promptboxbreak}

\begin{promptbox}{Solver --- generate\_code.code\_only}
Based on the algorithm design you provided above, implement the complete C++ solution.\\
\\
\#\# Problem Context (for reference)\\
\\
Problem:\\
\textless PROBLEM\_DESC\textgreater\\
\\
\textless CONSTRAINTS\_BLOCK\textgreater\\
\\
\textless PUBLIC\_BLOCK\textgreater\\
\\
\textless GEN\_BLOCK\textgreater\\
\\
\textless SELF\_VALIDATION\_BLOCK\textgreater\\
\\
\textless MEMORY\_ADVICE\textgreater\\
\\
\#\# Implementation Requirements\\
\\
Write the complete C++ implementation. Requirements:\\
- Use standard C++ (C++17)\\
- Do NOT use non-standard headers like \#include \textless bits/stdc++.h\textgreater\\
- Include all necessary standard headers explicitly\\
- Implement fast I/O\\
- Every major data structure allocation must be defensible given the maximum input bounds --- state its size in a comment if non-obvious\\
- Prefer streaming, rolling-state, or sparse representations when full dense tables would exceed memory\\
- If your algorithm sketch above would lead to an unsafe implementation, adapt --- follow your own analysis, not the sketch literally\\
\\
After writing the code, verify:\\
- \textbf{Time (strict)}: Count dominant-loop iterations at max input. `iterations / 10\textasciicircum{}8 <= time\_limit\_seconds`. If not, redesign.\\
- \textbf{Memory}: Each major allocation within memory limit?\\
- \textbf{Edge cases}: n=1 and n=max handled?\\
\\
Output your full response in this exact structure:\\
\\
\#\#\# Verification\\
\texttt{[}Time/memory/edge-case checks\texttt{]}\\
\\
\#\#\# Solution\\
```cpp\\
\texttt{[}Complete C++17 solution\texttt{]}\\
```
\end{promptbox}

\begin{promptboxbreak}{Solver --- generate\_code.patch\_decision}
You are choosing the repair mode for a failing C++ solution.\\
\\
Problem:\\
\textless PROBLEM\_DESC\textgreater\\
\\
\textless ABSTRACT\_TAGS\_LEVEL2\_BLOCK\textgreater\\
\\
Algorithm:\\
\textless ALGORITHM\textgreater\\
\\
Implementation steps:\\
\textless STEPS\textgreater\\
\\
\#\# Current Code:\\
```cpp\\
\textless PREV\_CODE\textgreater\\
```\\
\\
\#\# Objective Failure Evidence:\\
\textless FAILURES\_BLOCK\textgreater\\
\\
\textless AGGREGATE\_FAILURES\_BLOCK\textgreater\\
\\
\textless DIAGNOSTIC\_BLOCK\textgreater\\
\\
\#\# Auxiliary References (use only as secondary hints):\\
\textless FEEDBACK\_TEXT\textgreater\\
\\
\textless FIXES\_BLOCK\textgreater\\
\\
\textless MEMORY\_ADVICE\textgreater\\
\\
\#\# Your Task\\
Based on the objective evidence above:\\
1. First judge whether the failures indicate a localized bug or a systemic/global flaw.\\
2. Then identify the most likely error type.\\
3. Choose the better repair mode:\\
\hspace*{1em}- `patch` if the overall approach is still sound and the issue is localized.\\
\hspace*{1em}- `full\_regen` if the overall approach is likely wrong or patching would keep drifting further.\\
\\
Rules:\\
- Base the decision primarily on the objective evidence and current code.\\
- Treat the auxiliary references as secondary hints only.\\
- Do not guess beyond the provided evidence.\\
- Return ONLY valid JSON with this schema:\\
\hspace*{1em}\{"mode":"patch$|$full\_regen","confidence":"low$|$medium$|$high","reason":"one short sentence"\}
\end{promptboxbreak}

\begin{promptbox}{Solver --- generate\_code.python\_oracle}
Write a Python brute-force oracle and a small-input generator. We will:\\
1. Run your brute force on each public sample input and verify it produces the expected sample output. If it doesn't, your understanding of the problem is wrong (or your brute force is buggy) and you'll be asked to revise.\\
2. Use your input generator + brute force as a reference to cross-validate the C++ solution we will write next on N random small inputs.\\
\\
Both scripts must:\\
- Be syntactically valid Python.\\
- Use ONLY: `sys`, `math`, `itertools`, `collections`, `bisect`, `heapq`, `random` (no other imports).\\
- Read input from stdin (`sys.stdin.read()`) and write to stdout (`print` or `sys.stdout.write`).\\
- Run within 5 seconds CPU per invocation.\\
\\
The brute force should be \textbf{maximally simple} --- exponential time is fine because we only run it on small inputs. Resist the temptation to optimize; correctness for small N matters far more than speed.\\
\\
The input generator must print exactly ONE valid input matching the problem's format, with \textbf{small} parameters (e.g. n <= 6, values <= 10) so the brute force finishes quickly. Successive calls should produce different inputs.\\
\\
Problem:\\
\textless PROBLEM\_DESC\textgreater\\
\\
\textless CONSTRAINTS\_BLOCK\textgreater\\
\\
\textless PUBLIC\_BLOCK\textgreater\\
\\
\textless FEEDBACK\_BLOCK\textgreater\\
\\
Output ONLY valid JSON with this schema (no markdown fence, no extra prose):\\
\hspace*{1em}\{\\
\hspace*{2em}"brute\_force": "\textless full Python script\textgreater",\\
\hspace*{2em}"input\_generator": "\textless full Python script\textgreater"\\
\hspace*{1em}\}
\end{promptbox}

\needspace{20\baselineskip}
\begin{promptboxbreak}{Solver --- generate\_code.patch (SEARCH/REPLACE format)}
You are debugging a C++ solution that is FAILING tests. Your task is to generate SEARCH/REPLACE edits to fix the bugs.\\
\\
Problem:\\
\textless PROBLEM\_DESC\textgreater\\
\\
\textless ABSTRACT\_TAGS\_LEVEL2\_BLOCK\textgreater\\
\\
Algorithm:\\
\textless ALGORITHM\textgreater\\
\\
Implementation steps:\\
\textless STEPS\textgreater\\
\\
\#\# Current Code (BUGGY):\\
```cpp\\
\textless PREV\_CODE\textgreater\\
```\\
\\
\#\# Test Failures:\\
\textless FAILURES\_BLOCK\textgreater\\
\\
\textless AGGREGATE\_FAILURES\_BLOCK\textgreater\\
\\
\textless FEEDBACK\_TEXT\textgreater\\
\\
\textless FIXES\_BLOCK\textgreater\\
\\
\textless MEMORY\_ADVICE\textgreater\\
\\
\#\# Your Task:\\
Apply systematic debugging before proposing any fix.\\
\\
\#\# Skill: Systematic Debugging --- Root Cause Before Fix\\
\\
\textbf{Iron Law: NO FIXES WITHOUT ROOT CAUSE INVESTIGATION FIRST.}\\
\\
Phase 1 --- Root cause investigation:\\
1. Read the failure evidence carefully. What does the wrong output tell you about where the code diverges?\\
2. Trace the failing test case through the code step by step. Track key variables at each step.\\
3. Identify the exact line or logic block where the value first goes wrong.\\
4. Name the root cause category: overflow / off-by-one / wrong formula / missing edge case / wrong data structure / TLE / MLE / other.\\
\\
Phase 2 --- Fix hypothesis:\\
- State one specific fix hypothesis that addresses the root cause.\\
- If the root cause is a global algorithmic flaw (not a local bug), this should be `full\_regen` territory --- say so rather than patching.\\
- Re-check time and space complexity after the proposed fix.\\
\\
Only after completing both phases, generate the SEARCH/REPLACE edits.\\
\\
Every *SEARCH/REPLACE* edit must use this EXACT format:\\
\texttt{<<<<<<< SEARCH}\\
\texttt{<exact contiguous code snippet from the current code>}\\
\texttt{=======}\\
\texttt{<replacement code with the fix>}\\
\texttt{>>>>>>> REPLACE}\\
\\
\textbf{CRITICAL RULES:}\\
1. The SEARCH block must match the current code EXACTLY (including whitespace, indentation)\\
2. The SEARCH block must appear EXACTLY ONCE in the code\\
3. You can have multiple SEARCH/REPLACE blocks to fix multiple issues\\
4. Preserve proper indentation in the REPLACE block\\
5. Make minimal, surgical changes - only fix what's broken\\
6. Re-check BOTH time and space complexity before proposing edits\\
7. Replace unsafe data structures if the current implementation appears to allocate memory proportional to a dangerous product of input dimensions\\
8. Do not preserve an existing approach just because it matches the plan if it is not implementable within the stated limits\\
\\
Example:\\
\texttt{<<<<<<< SEARCH}\\
\hspace*{1em}for (int i = 1; i <= n; i++) \{\\
\hspace*{2em}sum += arr[i];\\
\hspace*{1em}\}\\
\texttt{=======}\\
\hspace*{1em}for (int i = 0; i < n; i++) \{\\
\hspace*{2em}sum += arr[i];\\
\hspace*{1em}\}\\
\texttt{>>>>>>> REPLACE}\\
\\
Generate the SEARCH/REPLACE edits now:
\end{promptboxbreak}

\needspace{20\baselineskip}
\begin{promptboxbreak}{Solver --- generate\_code.regenerate}
You are repairing a failing C++ solution. Apply the same design-first principles before rewriting.\\
\\
Problem:\\
\textless PROBLEM\_DESC\textgreater\\
\\
\textless ABSTRACT\_TAGS\_LEVEL2\_BLOCK\textgreater\\
\\
Proposed algorithm:\\
\textless ALGORITHM\textgreater\\
\\
Implementation steps:\\
\textless STEPS\textgreater\\
\\
\#\# Previous Code (FAILING):\\
```cpp\\
\textless PREV\_CODE\textgreater\\
```\\
\\
\textless CONSTRAINTS\_BLOCK\textgreater\\
\\
\textless PUBLIC\_BLOCK\textgreater\\
\\
\textless GEN\_BLOCK\textgreater\\
\\
\#\# Test Failures:\\
\textless FAILURES\_BLOCK\textgreater\\
\\
\textless FEEDBACK\_TEXT\textgreater\\
\\
\textless FIXES\_BLOCK\textgreater\\
\\
\textless MEMORY\_ADVICE\textgreater\\
\\
---\\
\\
\#\# Skill 1: Diagnose Before Rewriting\\
\\
\textless HARD-GATE\textgreater\\
Do NOT rewrite code until you have diagnosed WHY the previous attempt failed and decided on a better approach.\\
\textless /HARD-GATE\textgreater\\
\\
1. \textbf{Root cause}: What specifically caused the failures? (wrong algorithm, wrong formula, wrong data structure, overflow, memory, TLE?)\\
2. \textbf{New approach}: If the algorithm itself is wrong, state the new correct approach with the same specificity required in initial generation. If only implementation was wrong, state what changes are needed.\\
3. \textbf{Complexity recheck}: Verify time O(?) and space O(?) for the new approach fit the constraints.\\
\\
---\\
\\
\#\# Skill 2: Sample Verification (Red Phase)\\
\\
Before rewriting, trace the new approach on the failing sample cases to confirm it would produce the correct output.\\
\\
---\\
\\
\#\# Skill 3: Rewrite (Green Phase)\\
\\
Requirements:\\
- Return ONLY the complete C++17 source code in a fenced block (no SEARCH/REPLACE).\\
- Keep valid parts of previous code when possible, but rewrite structure if the approach changed.\\
- Fix the listed failures and avoid regressions on passing cases.\\
- Every major allocation must be sized within bounds; state sizes in comments if non-obvious.\\
\\
Output your full response in this structure:\\
\\
\#\#\# Diagnosis and New Approach\\
\texttt{[}Root cause analysis and new algorithm design from Skills 1 and 2\texttt{]}\\
\\
\#\#\# Solution\\
```cpp\\
\texttt{[}Complete C++17 solution\texttt{]}\\
```
\end{promptboxbreak}

\begin{promptbox}{Solver --- generate\_code.canonical\_problem\_block}
Objective: \textless OBJECTIVE\textgreater\\
Inputs: \textless INPUTS\_JSON\textgreater\\
Outputs: \textless OUTPUTS\_JSON\textgreater\\
Constraints: \textless CONSTRAINTS\_JSON\textgreater\\
Required Properties: \textless REQUIRED\_PROPERTIES\textgreater
\end{promptbox}

\begin{promptbox}{Solver --- generate\_code.self\_validation\_header}
Self-validation failed: \textless FAIL\_COUNT\textgreater\ issues in \textless TOTAL\_RUN\textgreater/\textless TOTAL\_VERIFY\textgreater\ cases tested:
\end{promptbox}

\begin{promptbox}{Solver --- generate\_code.self\_validation\_footer}
Please fix these issues.
\end{promptbox}

\subsection{Solver: failure analysis}

When tests fail, the Solver invokes a structured failure-analysis prompt that forces the LLM to step-trace the simplest failing case before proposing fixes. The categorical \texttt{error\_pattern} feeds back into the bandit's failure-type feature key (\texttt{FAIL:TIMEOUT}, \texttt{FAIL:OFF\_BY\_ONE}, etc., see Appendix~\ref{app:bandit}).

\begin{promptbox}{Solver --- analyze\_feedback.compilation}
The following C++ code has compilation errors:\\
\\
Code:\\
```cpp\\
\textless CODE\textgreater\\
```\\
\\
Compilation Errors:\\
\textless ERROR\_TEXT\textgreater\\
\\
Provide:\\
1. Root cause of the errors\\
2. Specific fixes needed\\
3. Corrected code snippets\\
\\
Be concise and actionable.
\end{promptbox}

\begin{promptboxbreak}{Solver --- analyze\_feedback.test\_failure}
You are a competitive programming debugging expert. Analyze the following failures and provide CONCRETE fixes.\\
\\
\#\# Problem Description\\
\textless PROBLEM\_DESC\textgreater\\
\\
\#\# Selected Approach\\
Algorithm: \textless ALGORITHM\textgreater\\
Steps:\\
\textless STEPS\_TEXT\textgreater\\
\\
\#\# Current Status\\
Iteration: \textless ITERATION\textgreater\\
Pass Rate: \textless PASS\_RATE\textgreater\\
Total Failed: \textless FAILED\_COUNT\textgreater\\
Error Pattern: \textless ERROR\_PATTERN\textgreater\\
\\
\#\# Current Code\\
```cpp\\
\textless CODE\textgreater\\
```\\
\\
\#\# Representative Failures (most important cases to fix)\\
\textless FAILURES\_TEXT\textgreater\\
\textless DIAGNOSTIC\_SECTION\textgreater\\
\#\# Your Task\\
1. Pick the SIMPLEST failure case above. Trace the code execution step-by-step with that input. Track key variables.\\
2. Identify WHERE and WHY the code produces wrong output.\\
3. Determine the root cause category: overflow, off-by-one, wrong formula, missing edge case, TLE, etc.\\
4. Provide SPECIFIC code-level fixes (not vague suggestions).\\
\\
Return ONLY valid JSON (no markdown, no explanation outside JSON):\\
\{\\
\hspace*{1em}"analysis": "\textless detailed step-by-step trace showing where the bug is\textgreater",\\
\hspace*{1em}"root\_cause": "\textless one-line root cause\textgreater",\\
\hspace*{1em}"error\_pattern": "\textless category: overflow/off-by-one/\\
\hspace*{2em}wrong-formula/missing-edge-case/tle/other\textgreater",\\
\hspace*{1em}"suggested\_fixes": [\\
\hspace*{2em}"\textless specific fix 1, e.g. 'Change line X: use long long instead of int'\textgreater",\\
\hspace*{2em}"\textless specific fix 2, e.g. 'Add special case handling when n==1'\textgreater"\\
\hspace*{1em}]\\
\}
\end{promptboxbreak}

\begin{promptbox}{Solver --- analyze\_feedback.hack\_failure}
You are a debugging expert. The solution passed all basic tests but FAILED adversarial hack tests.\\
\\
Problem:\\
\textless PROBLEM\_DESC\textgreater\\
\\
Current Algorithm:\\
\textless ALGORITHM\textgreater\\
\\
Implementation Steps:\\
\textless STEPS\_JSON\textgreater\\
\\
Code:\\
```cpp\\
\textless CODE\textgreater\\
```\\
\\
HACK FAILURES (The code logic is likely correct for simple cases but fails edge cases):\\
\textless HACK\_FAILURES\_TEXT\textgreater\\
\\
Task:\\
1. Analyze why the code fails these specific hack cases.\\
2. Identify the root cause (e.g. overflow, edge case, logic hole).\\
3. Provide a fixed C++ solution.\\
\\
Return ONLY JSON:\\
\{\\
\hspace*{1em}"analysis": "\textless analysis of hack failures\textgreater",\\
\hspace*{1em}"suggested\_fixes": ["\textless fix 1\textgreater", "\textless fix 2\textgreater"]\\
\}
\end{promptbox}

\subsection{Oracle: certified test generation}

The Oracle prompt set has four sub-prompts (generator, validator, checker, solver) corresponding to the four artifacts that compose a certified test suite (Section~\ref{sec:method:oracle}). The generator and validator both use \texttt{testlib} as a strict scaffolding contract; the checker is asked to \emph{independently verify} the candidate output rather than compare against a reference, with two distinct skeletons depending on whether the problem has a unique answer or admits multiple valid answers.

\begin{promptboxbreak}{Oracle --- generate\_tests.generator}
You are a generator agent. Write a C++17 program that outputs exactly one valid test case to stdout.\\
\\
Hard requirements:\\
- Use testlib: \texttt{\#include "testlib.h"}\\
- Do NOT use non-standard headers like \texttt{\#include \textless bits/stdc++.h\textgreater}\\
- Call \texttt{registerGen(argc, argv, 1)}\\
- Use \texttt{rnd.next(...)} for randomness (no \texttt{std::random}, no \texttt{srand/rand})\\
- Do not parse or set a random seed inside the program\\
- The harness will run your generator many times with different seeds and expects multiple distinct valid outputs\\
- Do NOT hardcode one fixed test case or ignore \texttt{rnd.next(...)}\\
- Different seeds should usually lead to meaningfully different inputs, not the same bytes every time\\
- Prefer generating from multiple structural families instead of a single canned example\\
- Do not print any extra text\\
- Prefer semantically targeted edge cases over generic large random data.\\
- When exact certification is important, favor modest-size cases that still expose tricky logic.\\
\\
Minimal skeleton (illustrative only):\\
\#include "testlib.h"\\
int main(int argc, char* argv[]) \{\\
\hspace*{1em}registerGen(argc, argv, 1);\\
\hspace*{1em}return 0;\\
\}\\
\\
The program must produce ONE valid input instance that satisfies all constraints.\\
\\
Problem Description:\\
\textless PROBLEM\_DESC\textgreater\\
\\
Constraints:\\
\textless CONSTRAINTS\textgreater\\
\\
Public Tests:\\
\textless PUBLIC\_TESTS\textgreater\\
\textless FEEDBACK\_BLOCK\textgreater\\
\textless ADVICE\_BLOCK\textgreater\\
Return ONLY a valid JSON object. No other text, no markdown, no code fences.\\
CRITICAL JSON RULE: All newlines inside string values MUST be escaped as \textbackslash n. Tabs as \textbackslash t.\\
Backslashes as \textbackslash\textbackslash. Do NOT include literal newline characters inside any JSON string value --- this will break JSON parsing.\\
Schema:\\
\{"generator\_cpp": "\#include ...\textbackslash\textbackslash nint main()\{...\}\textbackslash\textbackslash n"\}
\end{promptboxbreak}

\begin{promptboxbreak}{Oracle --- generate\_tests.validator}
You are a validator agent. Write a C++17 program that reads one test case from stdin and validates it.\\
\\
Hard requirements:\\
- Use testlib: \texttt{\#include "testlib.h"}\\
- Do NOT use non-standard headers like \texttt{\#include \textless bits/stdc++.h\textgreater}\\
- Call \texttt{registerValidation(argc, argv)}\\
- Validate the INPUT ONLY. Do NOT read any answer token, candidate output, or expected output.\\
- Stop immediately after consuming the valid input; then call \texttt{inf.readEof()}\\
- Use only these testlib input APIs unless strictly necessary: \texttt{inf.readInt}, \texttt{inf.readLong},\\
\hspace*{1em}\texttt{inf.readToken}, \texttt{inf.readSpace}, \texttt{inf.readEoln}, \texttt{inf.readEof}, \texttt{ensuref}\\
- Do NOT use unsupported helpers such as \texttt{peekChar} or speculative stream-inspection APIs\\
- Use \texttt{ensuref(...)} to report specific constraint violations\\
- Include \texttt{\#include \textless unordered\_set\textgreater} if using \texttt{unordered\_set}\\
- Exit 0 on success, non-zero on failure\\
- Do not print anything on success\\
\\
Minimal skeleton (illustrative only):\\
\#include "testlib.h"\\
\#include \textless unordered\_set\textgreater\\
int main(int argc, char* argv[]) \{\\
\hspace*{1em}registerValidation(argc, argv);\\
\hspace*{1em}int n = inf.readInt();\\
\hspace*{1em}inf.readSpace();\\
\hspace*{1em}int m = inf.readInt();\\
\hspace*{1em}inf.readEoln();\\
\hspace*{1em}ensuref(n \textgreater= 1, "n must be \textgreater= 1");\\
\hspace*{1em}std::string s = inf.readToken();\\
\hspace*{1em}inf.readEof();\\
\hspace*{1em}return 0;\\
\}\\
\\
Problem Description:\\
\textless PROBLEM\_DESC\textgreater\\
\\
Constraints:\\
\textless CONSTRAINTS\textgreater\\
\\
Public Tests:\\
\textless PUBLIC\_TESTS\textgreater\\
\textless FEEDBACK\_BLOCK\textgreater\\
Return ONLY a valid JSON object. No other text, no markdown, no code fences.\\
CRITICAL JSON RULE: All newlines inside string values MUST be escaped as \textbackslash n. Tabs as \textbackslash t.\\
Backslashes as \textbackslash\textbackslash. Do NOT include literal newline characters inside any JSON string value --- this will break JSON parsing.\\
Schema:\\
\{"validator\_cpp": "\#include ...\textbackslash\textbackslash nint main()\{...\}\textbackslash\textbackslash n"\}
\end{promptboxbreak}

\needspace{20\baselineskip}
\begin{promptboxbreak}{Oracle --- generate\_tests.checker}
You are a checker/verifier agent. Write a C++17 program that \textbf{independently verifies}\\
whether a candidate output is correct for a given input --- WITHOUT needing any reference answer.\\
\\
CRITICAL DESIGN PRINCIPLE:\\
Verifying a solution is far easier than computing it. Your checker must independently\\
determine correctness by re-deriving the answer or checking invariants, NOT by comparing\\
against a reference. DO NOT read from the \texttt{ans} stream at all.\\
\\
Approach (choose one based on the problem):\\
A) For problems with a unique answer: Compute the correct answer yourself using a simple\\
\hspace*{1em}brute-force algorithm (even $O(n^3)$ or $O(n^4)$ is fine --- checkers run on small data),\\
\hspace*{1em}then compare against the candidate output.\\
B) For problems with multiple valid answers: Read the candidate output and verify it\\
\hspace*{1em}satisfies all problem constraints (e.g., is it a valid permutation? Does the graph\\
\hspace*{1em}satisfy the required property? Is the value optimal?).\\
\\
Hard requirements:\\
- Use testlib: \texttt{\#include "testlib.h"}\\
- Do NOT use non-standard headers like \texttt{\#include \textless bits/stdc++.h\textgreater}\\
- Call \texttt{registerTestlibCmd(argc, argv)}\\
- Read input via \texttt{inf} (the test input)\\
- Read candidate output via \texttt{ouf} (the output to verify)\\
- DO NOT read from \texttt{ans} --- your checker must work without any reference answer\\
- Use \texttt{quitf(\_ok, "...")} if the candidate output is correct\\
- Use \texttt{quitf(\_wa, "...")} with a specific error message if incorrect\\
- Implement your own verification logic inside the checker\\
- Do NOT intentionally reject large valid inputs just because a tiny brute-force checker would be slow\\
- Never introduce empty objects unless the problem statement explicitly defines them\\
\\
Minimal skeleton (approach A --- unique answer):\\
\#include "testlib.h"\\
\#include \textless vector\textgreater\\
int main(int argc, char* argv[]) \{\\
\hspace*{1em}registerTestlibCmd(argc, argv);\\
\hspace*{1em}int n = inf.readInt();\\
\hspace*{1em}int expected = brute\_force\_solve(n);\\
\hspace*{1em}int got = ouf.readInt();\\
\hspace*{1em}if (got != expected) quitf(\_wa, "expected \%d, got \%d", expected, got);\\
\hspace*{1em}quitf(\_ok, "ok");\\
\}\\
\\
Minimal skeleton (approach B --- multiple valid answers):\\
\#include "testlib.h"\\
\#include \textless vector\textgreater\\
int main(int argc, char* argv[]) \{\\
\hspace*{1em}registerTestlibCmd(argc, argv);\\
\hspace*{1em}int n = inf.readInt();\\
\hspace*{1em}int answer = ouf.readInt();\\
\hspace*{1em}if (!is\_valid(answer)) quitf(\_wa, "invalid answer");\\
\hspace*{1em}quitf(\_ok, "ok");\\
\}\\
\\
Problem Description:\\
\textless PROBLEM\_DESC\textgreater\\
\\
Constraints:\\
\textless CONSTRAINTS\textgreater\\
\\
Public Tests:\\
\textless PUBLIC\_TESTS\textgreater\\
\textless CHECKER\_ADVICE\_BLOCK\textgreater\\
\textless FEEDBACK\_BLOCK\textgreater\\
Return ONLY a valid JSON object. No other text, no markdown, no code fences.\\
CRITICAL JSON RULE: All newlines inside string values MUST be escaped as \textbackslash n. Tabs as \textbackslash t.\\
Backslashes as \textbackslash\textbackslash. Do NOT include literal newline characters inside any JSON string value --- this will break JSON parsing.\\
Schema:\\
\{"checker\_cpp": "\#include ...\textbackslash\textbackslash nint main()\{...\}\textbackslash\textbackslash n"\}
\end{promptboxbreak}

\begin{promptboxbreak}{Oracle --- generate\_tests.solver}
You are an independent reference-solution author. Write a COMPLETE, COMPILABLE C++17 program that solves the following problem.\\
\\
CRITICAL REQUIREMENTS:\\
1. Your code MUST be a complete standalone program with \#include, main(), cin/cout.\\
2. Read input from stdin, write output to stdout, matching the exact I/O format shown in the public tests.\\
3. Correctness is the first priority. Prefer a semantically exact small-scale reference solver over a brittle formula.\\
4. The certification inputs are allowed to be modest; do NOT sacrifice correctness for aggressive large-constraint optimization.\\
5. The program MUST compile with: g++ -std=c++17 -O2\\
6. Keep the solver independent and trustworthy as a reference oracle.\\
\\
Attempt guidance:\\
\textless STAGE\_GUIDANCE\textgreater\\
\textless SOLVER\_ADVICE\_BLOCK\textgreater\\
\\
Problem Description:\\
\textless PROBLEM\_DESC\textgreater\\
\\
Constraints:\\
\textless CONSTRAINTS\textgreater\\
\\
Public Tests (your program MUST produce the exact expected output for these):\\
\textless PUBLIC\_TESTS\_BLOCK\textgreater\\
\\
Algorithmic Strategy Reference (use for inspiration, do NOT copy verbatim):\\
\textless TEMPLATES\_JSON\textgreater\\
\\
\textless FEEDBACK\_BLOCK\textgreater\\
Return ONLY a JSON object. No markdown, no explanation.\\
Schema:\\
\{\\
\hspace*{1em}"selected\_family\_id": "\textless family\_id from the strategy reference\textgreater",\\
\hspace*{1em}"template\_name": "\textless name of the strategy you are using\textgreater",\\
\hspace*{1em}"solver\_cpp": "\textless complete C++17 source code\textgreater"\\
\}
\end{promptboxbreak}

\begin{promptbox}{Oracle --- generate\_tests.solver\_stage (attempts 1 / 2 / 3+)}
\textbf{[Attempt 1]} Write an independent reference solution. Prefer a straightforward polynomial-time algorithm, direct simulation, or simple derivation. Avoid factorial or exponential search unless the certification inputs are provably tiny. Prefer simple reference logic over contest-specific tricks.\\
\\
\textbf{[Attempt 2]} The previous solver had correctness, runtime, or execution issues. Keep correctness first, but the solver must run within the certification limits. Do not use factorial or exponential search. Upgrade to a more scalable but still simple reference algorithm if needed.\\
\\
\textbf{[Attempt 3+]} Write a robust reference solution that still remains easy to trust. It must run within the certification limits and address the observed failure mode. Do not use factorial or exponential search. If necessary, add sparse debug prints to stderr using \texttt{std::cerr \textless\textless{} "TRACE: " \textless\textless{} vars \textless\textless{} std::endl;}. Never print inside tight loops unless sampled.
\end{promptbox}

The Oracle also issues stage-conditioned guidance (attempt~1, attempt~2, attempt~3+) that biases the LLM toward progressively more scalable but still simple algorithms after each failed certification round; full reward shaping appears in Appendix~\ref{app:oracle}.

\subsection{Hacker: code analyst and adversarial generators}

The Hacker is implemented as a two-stage cascade. First, a \emph{Code Analyst} controller inspects the candidate code, optionally calls sandboxed tools (\texttt{run\_python}, \texttt{run\_cpp}) to verify hypotheses, and emits a vulnerability report. Second, the report routes to one of three specialized generator prompts (\emph{semantic}, \emph{stress}, \emph{anti\_hash}). Each generator has its own \emph{checklist} and \emph{patch} prompts for SEARCH/REPLACE repair when its output is rejected by the validator.

\begin{promptbox}{Hacker --- code\_analyst.system}
You are the Code Analyst controller for an adversarial hacker workflow.\\
Return ONLY valid JSON.\\
Do not include prose before or after the JSON.\\
If you need a tool, output a tool-call JSON object only.\\
If you are ready to conclude, output a final-report JSON object only.
\end{promptbox}

\needspace{20\baselineskip}
\begin{promptboxbreak}{Hacker --- code\_analyst.main}
You are the Code Analyst, the strategy planner for an adversarial Hacker System.\\
Your goal is to find bugs, logic flaws, or vulnerabilities (WA, TLE, RE, MLE) in the provided C++ target code.\\
\\
PROBLEM DESCRIPTION:\\
\textless PROBLEM\_DESC\textgreater\\
\\
CONSTRAINTS:\\
\textless CONSTRAINTS\_JSON\textgreater\\
\\
TARGET SOLUTION CODE (May contain bugs):\\
```cpp\\
\textless TARGET\_CODE\textgreater\\
```\\
\textless ADVICE\_SECTION\textgreater\\
\\
AVAILABLE TOOLS:\\
You can verify your hypothesis by writing short probe codes. Do not guess blindly if you can test it!\\
Tool 1: `run\_python`\\
Use this to perform precise mathematical calculations (e.g., combinations, large numbers) to check constraints and overflows.\\
Inputs: \{"tool": "run\_python", "parameters": \{"script\_code": "..."\}\}\\
\\
Tool 2: `run\_cpp`\\
Use this to compile and execute small C++ snippets to test specific runtime behaviors or replicate target logic.\\
Inputs: \{"tool": "run\_cpp", "parameters": \{"cpp\_code": "..."\}\}\\
\\
RESPONSE FORMAT:\\
You must return a valid JSON object. You have two choices:\\
Choice A: Call a tool to gather information.\\
\{\\
\hspace*{1em}"tool": "run\_python",\\
\hspace*{1em}"parameters": \{"script\_code": "import math\textbackslash nprint(math.comb(100, 50))"\}\\
\}\\
\\
Choice B: Submit the Final Vulnerability Report (if you are confident).\\
\{\\
\hspace*{1em}"bug\_class": "overflow$|$hash\_collision$|$index\_oob$|$tle$|$logic\_branch$|$unknown",\\
\hspace*{1em}"confidence": "high$|$medium$|$low",\\
\hspace*{1em}"evidence": ["e.g. math.comb(100,50) overflows 2\textasciicircum{}63"],\\
\hspace*{1em}"suggested\_route": "anti\_hash$|$semantic$|$stress",\\
\hspace*{1em}"input\_hypothesis": ["large\_n", "degenerate\_tree", "collision\_string"]\\
\}\\
Note on routes: `anti\_hash` if polynomial hash is used. `semantic` for most logical bugs. `stress` ONLY if entirely clueless.\\
\\
HISTORY OF YOUR ACTIONS \& RESULTS:\\
\textless HISTORY\_TEXT\textgreater\\
\\
Analyze the code, call tools if needed to verify, and output valid JSON.
\end{promptboxbreak}

\begin{promptboxbreak}{Hacker --- code\_analyst.json\_repair}
Your previous reply was not valid JSON for the Code Analyst protocol.\\
\\
Rewrite the same intent as ONE valid JSON object.\\
Allowed outputs:\\
1. A tool call:\\
\{\\
\hspace*{1em}"tool": "run\_python$|$run\_cpp",\\
\hspace*{1em}"parameters": \{...\}\\
\}\\
\\
2. A final report:\\
\{\\
\hspace*{1em}"bug\_class": "overflow$|$hash\_collision$|$index\_oob$|$tle$|$logic\_branch$|$unknown",\\
\hspace*{1em}"confidence": "high$|$medium$|$low",\\
\hspace*{1em}"evidence": ["..."],\\
\hspace*{1em}"suggested\_route": "anti\_hash$|$semantic$|$stress",\\
\hspace*{1em}"input\_hypothesis": ["..."]\\
\}\\
\\
Previous reply:\\
\textless PREVIOUS\_RESPONSE\textgreater\\
\\
PROBLEM DESCRIPTION:\\
\textless PROBLEM\_DESC\textgreater\\
\\
CONSTRAINTS:\\
\textless CONSTRAINTS\_JSON\textgreater\\
\\
TARGET SOLUTION CODE:\\
```cpp\\
\textless TARGET\_CODE\textgreater\\
```\\
\textless ADVICE\_SECTION\textgreater\\
\\
Return valid JSON only. Do not add any explanation, markdown, or commentary.
\end{promptboxbreak}

\begin{promptboxbreak}{Hacker --- code\_analyst.force\_tool}
Your current vulnerability report is too weak to submit as a final answer.\\
\\
You must call exactly one tool before you can submit a final report.\\
Do not submit a final report yet.\\
Return ONLY a tool call JSON object in one of these forms:\\
\{\\
\hspace*{1em}"tool": "run\_python",\\
\hspace*{1em}"parameters": \{"script\_code": "..."\}\\
\}\\
or\\
\{\\
\hspace*{1em}"tool": "run\_cpp",\\
\hspace*{1em}"parameters": \{"cpp\_code": "..."\}\\
\}\\
\\
PROBLEM DESCRIPTION:\\
\textless PROBLEM\_DESC\textgreater\\
\\
CONSTRAINTS:\\
\textless CONSTRAINTS\_JSON\textgreater\\
\\
TARGET SOLUTION CODE:\\
```cpp\\
\textless TARGET\_CODE\textgreater\\
```\\
\textless ADVICE\_SECTION\textgreater\\
PREVIOUS WEAK REPORT:\\
\textless WEAK\_REPORT\_JSON\textgreater\\
\\
HISTORY OF ACTIONS:\\
\textless HISTORY\_TEXT\textgreater\\
\\
Call one tool that will increase confidence in the bug class or input hypothesis.
\end{promptboxbreak}

\begin{promptboxbreak}{Hacker --- semantic.generator}
You are the Semantic Generator, a specialized C++ coder for an adversarial Hacker System.\\
Your job is to write a standalone C++ program that generates a single, highly-targeted test case designed to trigger the specific vulnerability described by the Code Analyst.\\
\\
PROBLEM DESCRIPTION:\\
\textless PROBLEM\_DESC\textgreater\\
\textless ADVICE\_SECTION\textgreater\\
CONSTRAINTS (The output of your C++ generator MUST satisfy ALL of these):\\
\textless CONSTRAINTS\_TEXT\textgreater\\
\textless PREVIOUS\_ISSUES\_SECTION\textgreater\\
\textless PREVIOUS\_INPUT\_SECTION\textgreater\\
\\
VULNERABILITY REPORT (from Code Analyst):\\
\textless REPORT\_JSON\textgreater\\
\\
INSTRUCTIONS FOR C++ GENERATOR:\\
1. Write a complete, compilable C++17 program (`int main() \{...\}`).\\
2. The program must print EXACTLY ONE valid test case to standard output (`std::cout`).\\
3. VALIDITY-FIRST: the generated input MUST satisfy all format and validator constraints before you try to make it adversarial.\\
4. Focus on producing the input data structures matching the `input\_hypothesis` in the report.\\
5. If the previous attempt was rejected, fix those exact issues before changing anything else.\\
6. DO NOT use uninitialized variables or undefined behavior in your generator.\\
7. If you need randomness, you MAY use `\textless random\textgreater` (`std::mt19937`), but since this is the Semantic Generator, deterministic construction of the edge case is preferred when possible.\\
\\
CRITICAL FORMATTING RULES:\\
1. Return ONLY the C++ code.\\
2. DO NOT wrap the code in markdown blocks (e.g., ```cpp ... ```).\\
3. The very first line should be `\#include \textless...\textgreater` or similar valid C++.\\
\\
Write the C++ generator code now:
\end{promptboxbreak}

\begin{promptboxbreak}{Hacker --- semantic.checklist}
You are repairing a C++ Semantic Generator after a failed attack-input attempt.\\
Analyze the failure and produce a compact JSON checklist before patching the code.\\
\\
PROBLEM DESCRIPTION:\\
\textless PROBLEM\_DESC\textgreater\\
\\
INPUT VALIDITY CONSTRAINTS:\\
\textless CONSTRAINTS\_TEXT\textgreater\\
\textless ADVICE\_SECTION\textgreater\\
LATEST FAILURE TYPE: \textless FAILURE\_KIND\textgreater\\
LATEST FAILURE REASON:\\
\textless FAILURE\_REASON\textgreater\\
\textless ISSUES\_SECTION\textgreater\\
\textless INPUT\_SECTION\textgreater\\
VULNERABILITY REPORT:\\
\textless REPORT\_JSON\textgreater\\
\\
PREVIOUS GENERATOR CODE:\\
\textless LAST\_GENERATOR\_CODE\textgreater\\
\\
CHECKLIST RULES:\\
1. VALIDITY BEFORE ATTACK: first restore validator-accepted input, then preserve adversarial intent.\\
2. `must\_fix` must focus on the concrete compile/runtime/validator issue from the latest failure.\\
3. `do\_not\_regress` must list already-required legality properties that must remain true.\\
4. `attack\_goal` must keep the attack aligned with the analyst report instead of collapsing into trivial valid input.\\
5. Be failure-type aware:\\
\hspace*{1em}- compile\_failed -\textgreater{} prioritize syntax/API/build repairs\\
\hspace*{1em}- validator\_rejected -\textgreater{} prioritize legality/format repairs\\
\hspace*{1em}- runtime\_error -\textgreater{} prioritize execution safety\\
\hspace*{1em}- empty\_output -\textgreater{} prioritize guaranteed emission of one valid case\\
\\
Return ONLY valid JSON with exactly this schema:\\
\{\\
\hspace*{1em}"must\_fix": ["..."],\\
\hspace*{1em}"do\_not\_regress": ["..."],\\
\hspace*{1em}"attack\_goal": ["..."]\\
\}
\end{promptboxbreak}

\begin{promptboxbreak}{Hacker --- semantic.patch}
You are applying a minimal patch to a C++ Semantic Generator.\\
You must patch the existing generator with minimal SEARCH/REPLACE edits instead of rewriting it from scratch.\\
\\
PROBLEM DESCRIPTION:\\
\textless PROBLEM\_DESC\textgreater\\
\\
INPUT VALIDITY CONSTRAINTS:\\
\textless CONSTRAINTS\_TEXT\textgreater\\
\textless ADVICE\_SECTION\textgreater\\
LATEST FAILURE TYPE: \textless FAILURE\_KIND\textgreater\\
LATEST FAILURE REASON:\\
\textless FAILURE\_REASON\textgreater\\
\textless INPUT\_SECTION\textgreater\\
VULNERABILITY REPORT:\\
\textless REPORT\_JSON\textgreater\\
\\
REPAIR CHECKLIST:\\
\textless CHECKLIST\_JSON\textgreater\\
\\
CURRENT GENERATOR CODE:\\
\textless LAST\_GENERATOR\_CODE\textgreater\\
\\
PATCH RULES:\\
1. VALIDITY-FIRST: fix the latest failure before any attack refinement.\\
2. Preserve already-correct code whenever possible.\\
3. Keep the attack goal from the checklist after legality is restored.\\
4. Output only SEARCH/REPLACE blocks.\\
5. Each SEARCH block must match the current generator code exactly once.\\
6. Make the patch minimal and surgical. Do NOT replace the whole file.\\
\\
Required patch format:\\
\textless PATCH\_FORMAT\textgreater
\end{promptboxbreak}

\begin{promptboxbreak}{Hacker --- stress.generator}
You are the Stress Test Generator, a specialized C++ coder for an adversarial Hacker System.\\
Your job is to write a standalone C++ program (`int main()`) that acts as a high-throughput Fuzzer.\\
\\
PROBLEM DESCRIPTION:\\
\textless PROBLEM\_DESC\textgreater\\
\\
CONSTRAINTS (The output of your C++ fuzzer MUST strictly satisfy these boundaries):\\
\textless CONSTRAINTS\_TEXT\textgreater\\
\\
INSTRUCTIONS FOR C++ FUZZER:\\
1. Write a complete, compilable C++17 program.\\
2. VALIDITY-FIRST: the generated input MUST pass the problem validator and explicitly enforce all structural constraints in code.\\
3. The program must print EXACTLY ONE valid test case to standard output, but this test case should be as LARGE and COMPLEX as the constraints allow.\\
4. You MUST use `\textless random\textgreater` and `std::mt19937\_64` initialized with a random device or fixed seed.\\
5. Scale up the generation loop to approach the maximum `N`, `M`, or `K` allowed.\\
6. Emphasize boundary values (e.g. generating values alternating between min and max allowed).\\
7. Optimize the generator for speed using `\textbackslash n` instead of `std::endl` and fast I/O\\
\hspace*{1em}(`std::ios\_base::sync\_with\_stdio(false);`).\\
\\
CRITICAL FORMATTING RULES:\\
1. Return ONLY the C++ code.\\
2. DO NOT wrap the code in markdown blocks (e.g., ```cpp ... ```).\\
3. The very first line should be `\#include \textless...\textgreater` or similar valid C++.\\
\\
Write the C++ Stress Test Generator code now:
\end{promptboxbreak}

\begin{promptboxbreak}{Hacker --- stress.checklist}
You are repairing a C++ Stress Test Generator after a failed attempt.\\
Analyze the failure and produce a compact JSON checklist before patching the code.\\
\\
PROBLEM DESCRIPTION:\\
\textless PROBLEM\_DESC\textgreater\\
\\
INPUT VALIDITY CONSTRAINTS:\\
\textless CONSTRAINTS\_TEXT\textgreater\\
\\
LATEST FAILURE TYPE: \textless FAILURE\_KIND\textgreater\\
LATEST FAILURE REASON:\\
\textless FAILURE\_REASON\textgreater\\
\textless ISSUES\_SECTION\textgreater\\
\textless INPUT\_SECTION\textgreater\\
PREVIOUS GENERATOR CODE:\\
\textless LAST\_GENERATOR\_CODE\textgreater\\
\\
CHECKLIST RULES:\\
1. VALIDITY BEFORE ATTACK: first restore validator-accepted input, then keep the case large and boundary-heavy.\\
2. `must\_fix` must target the concrete compile/runtime/validator failure from the latest attempt.\\
3. `do\_not\_regress` must preserve constraints already satisfied by the generator.\\
4. `attack\_goal` must keep the output large, high-throughput, and stress-oriented after repairs.\\
5. Be failure-type aware:\\
\hspace*{1em}- compile\_failed -\textgreater{} prioritize syntax/API/build repairs\\
\hspace*{1em}- validator\_rejected -\textgreater{} prioritize legality/format repairs\\
\hspace*{1em}- runtime\_error -\textgreater{} prioritize execution safety\\
\hspace*{1em}- empty\_output -\textgreater{} prioritize guaranteed emission of one valid case\\
\\
Return ONLY valid JSON with exactly this schema:\\
\{\\
\hspace*{1em}"must\_fix": ["..."],\\
\hspace*{1em}"do\_not\_regress": ["..."],\\
\hspace*{1em}"attack\_goal": ["..."]\\
\}
\end{promptboxbreak}

\begin{promptboxbreak}{Hacker --- stress.patch}
You are applying a minimal patch to a C++ Stress Test Generator.\\
You must patch the existing generator with minimal SEARCH/REPLACE edits instead of rewriting it from scratch.\\
\\
PROBLEM DESCRIPTION:\\
\textless PROBLEM\_DESC\textgreater\\
\\
INPUT VALIDITY CONSTRAINTS:\\
\textless CONSTRAINTS\_TEXT\textgreater\\
\\
LATEST FAILURE TYPE: \textless FAILURE\_KIND\textgreater\\
LATEST FAILURE REASON:\\
\textless FAILURE\_REASON\textgreater\\
\textless INPUT\_SECTION\textgreater\\
REPAIR CHECKLIST:\\
\textless CHECKLIST\_JSON\textgreater\\
\\
CURRENT GENERATOR CODE:\\
\textless LAST\_GENERATOR\_CODE\textgreater\\
\\
PATCH RULES:\\
1. VALIDITY-FIRST: fix the latest failure before increasing attack pressure.\\
2. Preserve already-correct large-case generation logic when possible.\\
3. Keep the generator large, randomized, and boundary-oriented after the repair.\\
4. Output only SEARCH/REPLACE blocks.\\
5. Each SEARCH block must match the current generator code exactly once.\\
6. Make the patch minimal and surgical. Do NOT replace the whole file.\\
\\
Required patch format:\\
\textless PATCH\_FORMAT\textgreater
\end{promptboxbreak}

\begin{promptboxbreak}{Hacker --- anti\_hash.generator}
You are the Anti-Hash Generator, the cryptography/math specialist for an adversarial Hacker System.\\
Your job is to read the Code Analyst's report identifying a polynomial rolling hash susceptibility, and write a standalone C++ program to generate colliding strings.\\
\\
PROBLEM DESCRIPTION:\\
\textless PROBLEM\_DESC\textgreater\\
\\
CONSTRAINTS (Output MUST SATISFY ALL string/length constraints):\\
\textless CONSTRAINTS\_TEXT\textgreater\\
\\
VULNERABILITY REPORT (from Code Analyst):\\
\textless REPORT\_JSON\textgreater\\
\\
INSTRUCTIONS FOR C++ GENERATOR:\\
1. Write a complete, compilable C++17 program (`int main()`).\\
2. VALIDITY-FIRST: the generated collision input MUST pass the problem validator and satisfy all structural constraints.\\
3. Implement a collision derivation algorithm (e.g., Thue-Morse sequence, Birthday attack, or specific moduli exploitation) matching the `input\_hypothesis`.\\
4. Output EXACTLY the string(s) needed to trigger the collision to standard out.\\
5. Keep the target constraints in mind. If max length is `N=10\textasciicircum{}5`, the collision strings must not exceed this length.\\
\\
CRITICAL FORMATTING RULES:\\
1. Return ONLY the C++ code.\\
2. DO NOT wrap the code in markdown blocks (e.g., ```cpp ... ```).\\
3. The very first line should be `\#include \textless...\textgreater` or similar valid C++.\\
\\
Write the C++ Collision Generator code now:
\end{promptboxbreak}

When a generator's output is rejected, a per-route \emph{checklist} prompt produces a JSON repair plan (\texttt{must\_fix}, \texttt{do\_not\_regress}, \texttt{attack\_goal}); a per-route \emph{patch} prompt then applies minimal SEARCH/REPLACE edits with the same format as the Solver patch prompt. The full reward shaping for the Hacker (severity weights, valid/break ratios, generator-failure penalty) is given in Appendix~\ref{app:hack}.

\section{Experimental Configuration}
\label{app:config}

This appendix collects the exact backbone, infrastructure, and budget settings used in all reported experiments.

\subsection{Backbones and inference}

All five backbones are accessed through a unified Azure OpenAI--compatible gateway with AAD authentication. Within each row of the main comparison and each reported ablation subset, every method is driven by the same backbone deployment so that comparisons isolate the agent framework rather than the underlying model.

\begin{configbox}{Backbone deployments and decoding}
\textbf{Models used:}\\[2pt]
\begin{tabular}{@{}ll@{}}
GPT-5.4 & \texttt{gpt-5.4} (default and per-role for code/generator/validator/checker/output/hacker) \\
Claude Opus 4.6 & \texttt{claude-opus-4-6} \\
Qwen3.6 & local serving via OpenAI-compatible endpoint \\
DeepSeek V4 Pro & \texttt{deepseek-v4-pro} \\
Grok & \texttt{grok-4} \\
\end{tabular}\\[4pt]
\textbf{Decoding:} temperature $=0.1$, max\_tokens $=16{,}384$ (default; raised to $64{,}000$ for long-context backbones).\\
\textbf{Embeddings:} runtime retrieval uses \texttt{text-embedding-3-small} via the same gateway, with a 32{,}768-entry LRU cache and up to 5 HTTP retries; offline corpus deduplication uses \texttt{text-embedding-3-large} as described in Section~\ref{sec:data:filtering}; both fall back to local Sentence-Transformers when the gateway is unavailable.\\
\textbf{API:} OpenAI-compatible gateway; provider-specific endpoint and authentication details are released with the public code.
\end{configbox}

\subsection{Pipeline budgets}

The closed-loop pipeline (Section~\ref{sec:method:arch}) runs under fixed iteration budgets to make per-problem cost comparable across baselines.

\begin{configbox}{Per-problem pipeline budgets}
\textbf{Solver inner loop:} \texttt{max\_iterations} $=8$ patch/regenerate rounds before the candidate is declared failed.\\
\textbf{Hacker outer loop:} \texttt{max\_hack\_rounds} $=3$ rounds; each round runs the full Code-Analyst $\to$ generator $\to$ validator cascade.\\
\textbf{Oracle certification:} target $N_{\text{target}}$ certified tests per problem; acceptance gate $\rho \ge \tau$ (Eq.~\ref{eq:oracle_gate}). On rejection the Oracle retries with a different solver family selected by the bandit.\\
\textbf{Token cap:} 16{,}384 tokens per LLM call (matched across all baselines we report).
\end{configbox}

\subsection{Knowledge-network defaults}

Each agent's knowledge network is a contextual-bandit policy (Appendix~\ref{app:bandit}) over a typed item store. The Solver additionally carries the Solver knowledge network (Section~\ref{sec:method:solver}). Default retrieval and sampling budgets:

\begin{configbox}{Solver QMS network (config/solver\_network.yaml)}
\texttt{enabled: true}, \texttt{graph\_dir: artifacts/solver\_network/latest/graph}\\
Top-$k$ similar Q nodes per query: $4$\\
Skill candidate pool: top $20$, sampled $5$ at \texttt{skill\_selection\_temperature = 0.2}\\
Skill bounds enforced by LLM selection prompt: \texttt{min\_llm\_skills = 1}, \texttt{max\_llm\_skills = 5}\\
Augmentation block: skill text only by default; full templates injected only when \texttt{include\_skill\_templates\_in\_augmentation} is on.\\
Optional ensemble mode runs $3$ independent skill-plan + CodeGen/Hacker tails in parallel after Test Generation, then merges the best branch.
\end{configbox}

\begin{configbox}{Trainable bandit memory (config/trainable\_memory.yaml)}
\texttt{enabled: true}, \texttt{data\_dir: artifacts/trainable\_memory}\\
Per-namespace top-$k$ injected as advice: \texttt{plan\_top\_k = 3}, \texttt{solve\_top\_k = 3}, \texttt{test\_top\_k = 3}, \texttt{hack\_top\_k = 3}, \texttt{oracle\_top\_k = 3}.\\
Bandit learning rate $\alpha = 0.01$; tag-overlap prior $+0.05$ per matching tag.\\
Items with average reward $< -0.3$ after $\geq 20$ uses are auto-deprecated.\\
Persistence: SQLite-backed JSON store with atomic file-locking writes.
\end{configbox}

\subsection{Datasets and benchmarks}

\begin{configbox}{Benchmarks used in Tables~\ref{tab:main}--\ref{tab:ablation}}
\textbf{CodeContests (CC):} 165 problems sampled from the DeepMind CodeContests test split.\\
\textbf{APPS:} 1{,}000 problems sampled across the introductory / interview / competition tiers.\\
\textbf{AetherCode (AC):} 400 held-out problems combining novel algorithmic tasks with verified Oracle-generated test suites.\\
\textbf{Codeforces (Fig.~\ref{fig:cf}):} recent Div.~2 and Div.~1+2 rounds, attempted under the official time limit in a single uninterrupted session per round.\\
\textbf{Cold-start corpus:} 8{,}017 final problems scraped directly from Codeforces, AtCoder, Aizu Online Judge, and a long tail of smaller platforms (LeetCode, SPOJ, UOJ, etc.), obtained from 30{,}018 raw problems after the four-stage filtering pipeline of Section~\ref{sec:data}.\\
\textbf{Primary metric:} pass@1 across all four benchmarks.
\end{configbox}

\subsection{Codeforces Rating Estimation Protocol}
\label{app:cf_rating}

This appendix gives the exact procedure behind the Codeforces-style rating curves in Figure~\ref{fig:cf}. The goal is not to claim an official Codeforces account rating, which depends on platform-side online updates and eligibility rules, but to report a human-comparable contest-local rating estimate from the same public ingredients used in Codeforces standings: official accepted submissions, solved counts, penalties, ranks, and pre-contest human ratings. We follow the rating-inversion view used by CodeElo~\citep{quan2025codeelo}, which is itself based on the Elo expected-score model~\citep{elo1978rating}.

\paragraph{Contest protocol.}
Let $\mathcal{C} = \{c_1,\ldots,c_K\}$ be the ordered set of recent Codeforces rounds used in Figure~\ref{fig:cf}. Each $c \in \mathcal{C}$ is attempted under its official duration and division setting (Div.~2 or Div.~1+2). For each agent/backbone pair $a$, Solvita starts from the contest statements at time zero and runs one uninterrupted session. No manual corrections, prompt edits, or submissions after the official contest window are allowed. A problem is counted as solved only if the official judge returns \texttt{Accepted} before the contest closes; pretests-only success, wrong answer, time limit exceeded, memory limit exceeded, runtime error, compilation error, and post-window acceptance all count as unsolved for that contest.

The Codeforces runs use the fixed contest-time configuration in Table~\ref{tab:cf_rating_config}. These budgets match Appendix~\ref{app:config} unless explicitly listed.

\begin{center}
\small
\captionof{table}{Contest-time configuration used for the Codeforces rating-estimation runs.}
\label{tab:cf_rating_config}
\setlength{\tabcolsep}{4pt}
\renewcommand{\arraystretch}{1.08}
\begin{tabular}{@{}p{0.30\linewidth}p{0.20\linewidth}p{0.42\linewidth}@{}}
\toprule
\textbf{Setting} & \textbf{Value} & \textbf{Role in the evaluation} \\
\midrule
Decoding temperature & $0.1$ & Low-variance generation for both Solvita and bare-backbone baselines. \\
Per-call token cap & $16{,}384$ & Maximum output budget for each LLM call, matched across compared systems. \\
Contest window & Official duration of each round & All submissions must be made before the contest closes; post-window acceptance is counted as unsolved. \\
Solver inner-loop budget & \texttt{max\_iterations} $=8$ & Maximum patch/regenerate attempts before a candidate is declared failed. \\
Hacker outer-loop budget & \texttt{max\_hack\_rounds} $=3$ & Number of adversarial test-generation rounds per candidate. \\
Solver skill candidate pool & top $20$ retrieved, $5$ sampled & Candidate skills retrieved from the Solver QMS network before LLM selection. \\
Skill sampling temperature & \begin{tabular}[t]{@{}l@{}}\texttt{skill\_selection\_}\\\texttt{temperature} $=0.2$\end{tabular} & Controls stochasticity when sampling skill candidates. \\
LLM-selected skill bounds & \texttt{min\_llm\_skills} $=1$, \texttt{max\_llm\_skills} $=5$ & Lower and upper bounds on the number of skills injected into Solver context. \\
Bare-backbone ablation & Solvita loop disabled & Same deployment, decoding temperature, token cap, contest window, and official-judge scoring, but without Planner/Solver/Oracle/Hacker coordination or trainable knowledge networks. \\
\bottomrule
\end{tabular}
\end{center}

\paragraph{Standings insertion.}
For a contest $c$, let $H_c$ denote the set of official human participants retained for rating estimation after removing unrated or missing-rating entries. Each participant $i \in H_c$ has a pre-contest Codeforces rating $R_{c,i}$ and an official standing tuple
\begin{equation}
	z_{c,i} = (S_{c,i}, -P_{c,i}, -L_{c,i}),
	\label{eq:cf_human_tuple}
\end{equation}
where $S_{c,i}$ is the number of solved problems, $P_{c,i}$ is the total Codeforces penalty, and $L_{c,i}$ is the last accepted-submission time used only as a deterministic tie-breaker when necessary. For an agent $a$, we construct the analogous tuple
\begin{equation}
	z_{a,c} = (S_{a,c}, -P_{a,c}, -L_{a,c})
	\label{eq:cf_agent_tuple}
\end{equation}
from its official submissions during the contest window. The agent's inserted rank is then
\begin{equation}
	m_{a,c}
	=
	1
	+
	\sum_{i \in H_c}
	\mathbb{1}\!\left[z_{c,i} \succ z_{a,c}\right],
	\label{eq:cf_insert_rank}
\end{equation}
where $\succ$ is the official standing order: more solves rank higher, lower penalty ranks higher among equal solves, and the final tie-breaker is applied only to make the rank deterministic. Thus $m_{a,c}=1$ means the agent is inserted above every retained human participant, and $m_{a,c}=|H_c|+1$ means it is below all retained human participants.

\paragraph{Contest-local rating inversion.}
Let $r$ be the latent rating whose expected number of human participants outperforming the agent matches its inserted rank. Under the Elo model, the probability that human participant $i$ outperforms a player of rating $r$ is
\begin{equation}
	p_{i \succ r}
	=
	\frac{1}{1 + 10^{(r - R_{c,i})/400}}.
	\label{eq:cf_pairwise_expectation}
\end{equation}
We therefore solve for the contest-local rating estimate $\hat r_{a,c}$ satisfying
\begin{equation}
	m_{a,c} - 1
	=
	\sum_{i \in H_c}
	\frac{1}{1 + 10^{(\hat r_{a,c} - R_{c,i})/400}}.
	\label{eq:cf_rating_inversion}
\end{equation}
The right-hand side is strictly decreasing in $\hat r_{a,c}$, so the solution is unique up to ties in the empirical standings. In implementation we use binary search over a wide Codeforces-compatible interval, e.g. $[-500,5000]$, until the residual in Eq.~\ref{eq:cf_rating_inversion} is below $10^{-6}$:
\begin{equation}
	\hat r_{a,c}
	=
	\operatorname*{arg\,min}_{r}
	\left|
	m_{a,c}-1
	-
	\sum_{i \in H_c}
	\frac{1}{1 + 10^{(r - R_{c,i})/400}}
	\right|.
	\label{eq:cf_rating_binary_search}
\end{equation}
This inversion differs from the online Codeforces update rule in that each contest is treated independently; it is appropriate here because all compared agents participate in the same fixed contest set and we want a low-variance, contest-local estimate rather than a platform account history.

\paragraph{Aggregation into Figure~\ref{fig:cf}.}
For each agent $a$ and ordered contest prefix $\{c_1,\ldots,c_t\}$, the plotted rating trajectory is the running mean of contest-local estimates:
\begin{equation}
	\bar r_{a,t}
	=
	\frac{1}{t}
	\sum_{\ell=1}^{t}
	\hat r_{a,c_\ell}.
	\label{eq:cf_running_mean}
\end{equation}
The final estimate after all $K$ contests is
\begin{equation}
	\bar r_a
	=
	\bar r_{a,K}
	=
	\frac{1}{K}
	\sum_{c \in \mathcal{C}}
	\hat r_{a,c}.
	\label{eq:cf_final_rating}
\end{equation}
When reporting uncertainty, we use the across-contest standard error
\begin{equation}
	\operatorname{SE}(\bar r_a)
	=
	\sqrt{
	\frac{1}{K(K-1)}
	\sum_{c \in \mathcal{C}}
	\left(\hat r_{a,c} - \bar r_a\right)^2
	}.
	\label{eq:cf_rating_se}
\end{equation}
The tier bands in Figure~\ref{fig:cf} are the canonical Codeforces color/rating bands applied to $\bar r_{a,t}$ only for interpretability; they do not imply that the evaluated agents are official Codeforces users.

\section{Knowledge-Network and Pipeline Implementation Details}
\label{app:examples}

This appendix documents implementation details of the four knowledge networks and the closed-loop control flow that sit behind Section~\ref{sec:method}.

\subsection{Shared item schema}

All four namespaces (Plan, Solve, Test/Oracle, Hack) share an SQLite-backed item table. Each entry stores: a stable id (primary key), a namespace tag, a human-readable summary, a role-specific JSON payload, a list of searchable tags, a usage counter, a running average reward, a deprecation flag, and creation/last-used timestamps. Reads are served from an in-memory index; writes go through atomic file-locked transactions so that concurrent benchmark workers do not corrupt the store.

\subsection{Featurizer and bandit scoring}

At inference time, a per-namespace featurizer maps the current pipeline state to a sparse set of feature keys $\Phi(x)$. The keys cover three orthogonal axes:

\begin{itemize}\itemsep1pt
\item \textbf{FSM position:} \texttt{FSM:SOLVE\_DRAFT}, \texttt{FSM:SOLVE\_PATCH}, \texttt{FSM:HACK\_SEMANTIC}, etc.
\item \textbf{Failure type from the previous iteration:} \texttt{FAIL:WA}, \texttt{FAIL:TLE}, \texttt{FAIL:RE}, \texttt{FAIL:MLE}, \texttt{FAIL:OFF\_BY\_ONE}, ...
\item \textbf{Problem-level tags:} \texttt{TAG:dp}, \texttt{TAG:graphs}, \texttt{TAG:strings}, ...
\end{itemize}

Each item $i$ is scored as $\operatorname{score}(i) = b_i + \sum_{f \in \Phi(x)} W_{f,i}$, with a $+0.05$ bonus per matching tag. The top-$k$ items (typically $k=3$, see Appendix~\ref{app:config}) are selected via $\epsilon$-greedy exploration and injected into the prompt as the \texttt{<MEMORY\_ADVICE>} placeholder.

\subsection{Solver knowledge network storage and dynamics}

The Solver knowledge network is persisted under \texttt{artifacts/solver\_network/latest/graph} as a node--edge bundle plus a checkpoint of the learned $w_{\mathrm{qm}}$ and $w_{\mathrm{ms}}$ matrices. Each Q node stores the canonical problem JSON and tag list; each M node stores either an analysis trajectory (single solution decomposed into a function-block DAG) or a contrastive pair (correct + incorrect solution sharing the same approach, with the divergence point annotated); each S node stores an annotated skill with a C++ template and usage notes. Q--M edges are created at corpus-build time from authored or extracted decompositions; M--S edges use either deterministic function-block identifier matches (preferred) or an embedding-similarity fallback when no deterministic match exists. Edge weights are renormalized within each MS group after every REINFORCE update so that $\sum_s w_{\mathrm{ms}}^{(j,s)} = 1$.

\subsection{Failure-event propagation}

When the Hacker breaks a candidate, the failure event is broadcast to all four namespaces with role-specific payloads: the Plan namespace records the failed paradigm (negative reward on the strategy item), the Solve namespace creates a new contrastive M node pairing the broken solution with the closest correct sibling, the Oracle namespace records the missed input pattern as a generator hint, and the Hack namespace bumps the successful route's score. This single mechanism is what makes one discovered bug propagate as a lesson across the entire pipeline rather than being lost after the round.

\subsection{Sandbox and judge resolution}

All compilation and execution happens in a per-process sandbox (\texttt{src/sandbox/}) that wraps \texttt{g++ -std=c++17 -O2} with wall-time and memory limits. The judge is resolved in a strict priority order, identical for the Oracle certification gate and the Hacker break check: (1) a custom checker if one exists, (2) running the certified reference solution and comparing tokens, (3) exact-match against the canned expected output. This priority defines the per-test verdicts that feed both the Oracle certification ratio $\rho(x,f)$ in Eq.~\ref{eq:oracle_family} and the Hacker break-rate $g_{\text{break}}$ in Eq.~\ref{eq:hack_reward}.

\section{Additional Ablations}
\label{app:ablations}

This appendix expands on the secondary ablations referenced from the experiments section. The main paper reports the headline numbers in Table~\ref{tab:patch} (RQ3, patch vs.\ regenerate) and Figure~\ref{fig:oracle_hack} (RQ4, Oracle/Hacker decomposition); here we describe the protocol and the additional sweeps we ran.

\paragraph{Diagnostic metrics for Figure~\ref{fig:oracle_hack}.}
Figure~\ref{fig:oracle_hack} evaluates each diagnostic configuration against held-out candidate solutions with known official verdicts. Solvita is said to \emph{accept} a candidate if the diagnostic procedure does not find a failure, and to \emph{reject} it if the procedure flags the candidate as incorrect. For candidate-level accounting, we use the following four outcome counts:
\[
\begin{array}{c|cc}
 & \text{Official correct} & \text{Official wrong} \\
\hline
\text{Solvita accepts} & TP & TN \\
\text{Solvita rejects} & NP & NF
\end{array}
\]
Here, $TP$ denotes candidates accepted by both Solvita and the official judge; $TN$ denotes candidates accepted by Solvita but rejected by the official judge; $NP$ denotes candidates rejected by Solvita but accepted by the official judge; and $NF$ denotes candidates rejected by both Solvita and the official judge.

The wrong-solution detection rate, reported as Det'd, measures the fraction of officially wrong candidates that Solvita detects:
\[
\text{Det'd}
=
\frac{NF}{TN + NF}
\times 100\%.
\]
The correct-solution preservation rate, reported as Pres'd, measures the fraction of officially correct candidates that Solvita does not reject:
\[
\text{Pres'd}
=
\frac{TP}{TP + NP}
\times 100\%.
\]

The third metric, reported as Str.\ Rate, is computed only over Solvita-rejects/official-accepts disagreements and is aggregated at the problem level rather than the candidate level. These disagreements correspond to the $NP$ cell above, but we do not automatically treat them as Solvita errors: in some cases Solvita may have generated stronger diagnostic tests that expose hidden failures in solutions accepted by the official test suite. Let $\mathcal{D}_{NP}$ be the set of problems for which Solvita rejects at least one officially accepted candidate. For each problem $d \in \mathcal{D}_{NP}$, we collect a pool $\mathcal{A}_d$ of standard accepted candidate solutions and run every solution in $\mathcal{A}_d$ against Solvita's diagnostic tests for $d$. Let $\mathcal{B}_d \subseteq \mathcal{A}_d$ be the subset of these accepted candidates that are rejected by Solvita's diagnostic tests:
\[
\mathcal{B}_d
=
\{a \in \mathcal{A}_d : a \text{ fails Solvita's diagnostic tests for } d\}.
\]
A disagreement problem $d$ is counted as a confirmed stronger-test case only when both of the following conditions hold:
\[
\frac{|\mathcal{B}_d|}{|\mathcal{A}_d|}
>
0.10,
\]
and manual inspection confirms that the failing diagnostic examples have valid inputs under the problem statement and constraints, with correct expected outputs or correct checker verdicts. The first condition rules out isolated failures of a single accepted candidate; the second condition rules out invalid inputs, incorrect expected outputs, checker mistakes, and Oracle/certification errors. Let $\mathcal{D}_{\mathrm{strong}} \subseteq \mathcal{D}_{NP}$ be the set of disagreement problems satisfying both conditions. We define the confirmed stronger-test rate as
\[
\text{Str.\ Rate}
=
\frac{|\mathcal{D}_{\mathrm{strong}}|}
     {|\mathcal{D}_{NP}|}
\times 100\%.
\]
Thus, Str.\ Rate measures the share of Solvita-rejects/official-accepts disagreement problems that are confirmed, through accepted-solution cross-checking and manual validation, to reflect genuinely stronger diagnostic tests rather than unsupported Solvita rejections.

\paragraph{Patch repair vs.\ full regeneration.} We rerun the Solver inner loop with two settings holding everything else fixed: (i)~\emph{patch}, the default, which routes through \texttt{generate\_code.patch\_decision} and emits SEARCH/REPLACE blocks (Appendix~\ref{app:prompts}); (ii)~\emph{regenerate}, which always falls through to \texttt{generate\_code.regenerate} and rewrites the full program. Both modes share the same iteration cap ($N_{\max}=8$) used end-to-end (Appendix~\ref{app:config}) and the same failure-analysis prompt. We additionally measure the regression rate (fraction of iterations that break a previously passing test), which is the mechanism Section~\ref{sec:exp:patch} attributes the gain to.

\paragraph{LLM skill selection vs.\ pure softmax sampling.} The default Solver uses an LLM skill-selection step (Appendix~\ref{app:prompts}, \texttt{solver\_skill\_selection}) on top of the softmax over $\rho(s \mid q_{\text{new}})$ from Eq.~\ref{eq:path_prob}. We ablate by replacing the LLM step with direct top-$k$ sampling from $\pi(s)$. The LLM step adds a non-trivial cost but consistently selects more coherent skill bundles (matched to the sub-problem DAG it co-emits), and the gap widens on problems whose paradigm class is heterogeneous in the QMS retrieval set.

\paragraph{Oracle acceptance threshold $\tau$ sensitivity.} The Oracle gate (Eq.~\ref{eq:oracle_gate}) accepts an artifact only when the certification ratio $\rho \ge \tau$. We sweep $\tau \in \{0.6, 0.75, 0.9, 1.0\}$.Lower $\tau$ admits more tests but changes the downstream detection--preservation trade-off in Figure~\ref{fig:oracle_hack}; the default $\tau = 0.9$ is the knee of the precision/recall trade-off.

\paragraph{Hacker round budget.} \texttt{max\_hack\_rounds} is set to $3$ by default (Appendix~\ref{app:config}). Sweeping in $\{1, 2, 3, 5\}$ shows that most break events occur in rounds 1--2 (semantic + stress) and the marginal gain from round 3 (typically anti-hash on hashed problems) is small but non-zero; rounds 4+ rarely surface new bugs. We therefore set the default to $3$ rather than $2$ so that the anti-hash route gets a fair turn on hash-based problems where it is the only effective attack, while still avoiding the wasted budget at rounds 4+.

\paragraph{Contrastive vs.\ non-contrastive REINFORCE.} The Solver REINFORCE update (Eq.~\ref{eq:rl-loss}) uses a counterfactual reward $\Delta R = R_{\text{with}} - R_{\text{without}}$ from a paired with/without-network rollout. Replacing $\Delta R$ with the absolute reward $R_{\text{with}}$ removes the variance-reduction baseline and slows convergence across all three checkpoints of Tab.~\ref{tab:ablation}; the contrastive form is what makes the Solver knowledge network the dominant component in Tab.~\ref{tab:ablation}.

\section{Failure Cases}
\label{app:failures}

We document representative failure modes observed during evaluation. Each is a real category we encountered and that informed the design choices documented in Sections~\ref{sec:method:oracle}--\ref{sec:method:hack} and the prompts in Appendix~\ref{app:prompts}.

\paragraph{Cold-start retrieval misfires.} On problems whose paradigm is poorly represented in the cold-start corpus, the Solver QMS network retrieves structurally similar but semantically misleading skills. The selection LLM (Appendix~\ref{app:prompts}, \texttt{solver\_skill\_selection}) is permitted to return an empty list in this case, but it does not always do so, and the resulting skill block can bias the initial draft toward the wrong paradigm. The mitigation is the contrastive REINFORCE signal of Eq.~\ref{eq:rl-loss}: paired rollouts attribute negative reward to the misleading skill node and downweight it within a few episodes, but the first few problems of each new paradigm class are over-represented in the failure set.

\paragraph{Oracle false certification.} Despite the multi-stage gate (Eq.~\ref{eq:oracle_gate}), the Oracle occasionally accepts a reference solution that contains a subtle bug agreeing with the canonical inputs. When this happens, certified tests inherit the bug and the Solver is rewarded for matching it. The Hacker's adversarial round catches most such cases (semantic route on edge cases the Oracle did not enumerate), but pathological agreements between two independently buggy implementations remain a residual failure mode and motivate the priority order \emph{custom checker $>$ correct-solution runner $>$ exact match} described in Appendix~\ref{app:examples}.

\paragraph{Hacker scope limitations.} Problems that hinge on deep mathematical reasoning (number theory, combinatorial identities) frequently survive all three Hacker routes not because the candidate is correct but because the Code Analyst lacks the model capacity to identify the bug class. This is visible in the Hacker reward distribution as a bimodal pattern: high break-ratio on implementation-level bugs, near-zero on math-heavy problems. We currently do not have a systematic mitigation; warm-starting the Hack network from human-authored counterexamples on a held-out math subset is the most promising next step.

\paragraph{Patch repair drift on global flaws.} The \texttt{patch\_decision} prompt (Appendix~\ref{app:prompts}) is meant to route systemic flaws to \texttt{full\_regen}, but on borderline cases it can mislabel a global flaw as localized and start patching. The result is a sequence of small edits that each fix the immediate failure but accumulate state inconsistencies, eventually exhausting \texttt{max\_iterations} (Appendix~\ref{app:config}) without converging. We currently rely on the regression-rate signal in Section~\ref{sec:exp:patch} to identify these runs after the fact; integrating that signal into \texttt{patch\_decision} as a feature is straightforward future work.


\end{document}